\documentclass[10pt,journal,cspaper,compsoc]{IEEEtran}
\pdfoutput=1

\usepackage[nocompress]{cite}
\usepackage[pdftex]{graphicx}
\graphicspath{{./images/}{./images2/}{./biography/}{./figures/}}
\DeclareGraphicsExtensions{.pdf,.jpeg,.png}
\usepackage[cmex10]{amsmath}
\usepackage{algorithm}
\usepackage{algorithmic}
\usepackage[tight,normalsize,sf,SF]{subfigure}
\usepackage{url}
\usepackage[cmex10]{amsmath}
\usepackage{amssymb}
\usepackage{stfloats}
\usepackage{amsthm}
\usepackage{bbm}
\usepackage{relsize}
\usepackage{booktabs}
\usepackage{multirow}
\usepackage[T1]{fontenc}
\usepackage{textcomp}

\newcommand{\tensor}[1]{\boldsymbol{\mathcal{#1}}}

\newcommand{\condp}[2]{p\left(#1 \middle\vert #2 \right)}

\newcommand{\normalpdf}[3]{\mathcal{N}\left(#1 \middle\vert#2 , #3\right)}

\newcommand{\hadamard}{\operatornamewithlimits{\mathlarger{\mathlarger{\mathlarger{\circledast}}}}}

\theoremstyle{plain}
\newtheorem{theorem}{Theorem}[section]

\theoremstyle{definition}

\theoremstyle{remark}

\begin{document}

\title{Bayesian CP Factorization of Incomplete Tensors with Automatic Rank Determination}

\author{Qibin Zhao,~\IEEEmembership{Member,~IEEE,}
        Liqing Zhang,~\IEEEmembership{Member,~IEEE,} and
        Andrzej Cichocki~\IEEEmembership{Fellow,~IEEE} 
\IEEEcompsocitemizethanks{\IEEEcompsocthanksitem Q. Zhao is at the Laboratory for Advanced Brain Signal Processing, RIKEN Brain Science Institute, Japan and the Department of Computer Science and Engineering, Shanghai Jiao Tong University, China.
\IEEEcompsocthanksitem L. Zhang is at the MOE-Microsoft Laboratory for Intelligent Computing and Intelligent Systems and the Department of Computer Science and Engineering, Shanghai Jiao Tong University, China.
\IEEEcompsocthanksitem A. Cichocki is at the Laboratory for Advanced Brain Signal Processing,
RIKEN Brain Science Institute, Japan and the Systems Research Institute at the Polish Academy of Science,  Warsaw,  Poland.

}
\thanks{} }

\IEEEcompsoctitleabstractindextext{%

\begin{abstract}
CANDECOMP/PARAFAC (CP) tensor factorization of incomplete data is a powerful technique for tensor completion through explicitly capturing the multilinear latent factors. The existing CP algorithms require the tensor rank to be manually specified, however, the determination of tensor rank remains a challenging problem especially for \emph{CP rank}. In addition, existing approaches do not take into account uncertainty information of latent factors, as well as missing entries. To address these issues, we formulate CP factorization using a hierarchical probabilistic model and employ a fully Bayesian treatment by incorporating a sparsity-inducing prior over multiple latent factors and the appropriate hyperpriors over all hyperparameters, resulting in automatic rank determination. To learn the model, we develop an efficient deterministic Bayesian inference algorithm, which scales linearly with data size. Our method is characterized as a tuning parameter-free approach, which can effectively infer underlying multilinear factors with a low-rank constraint, while also providing predictive distributions over missing entries. Extensive simulations on synthetic data illustrate the intrinsic capability of our method to recover the ground-truth of \emph{CP rank} and prevent the overfitting problem, even when a  large amount of entries are missing. Moreover, the results from real-world applications, including image inpainting and facial image synthesis, demonstrate that our method outperforms state-of-the-art approaches for both tensor factorization and tensor completion in terms of predictive performance.
\end{abstract}

}

\maketitle

\IEEEdisplaynotcompsoctitleabstractindextext

\IEEEpeerreviewmaketitle

\section{Introduction}
\IEEEPARstart{T}{ensors} (i.e., multiway arrays) provide an effective and faithful representation of the structural properties of data, in particular, when multidimensional data or a data ensemble affected by multiple factors are involved. For instance, a video sequence can be represented by a third-order tensor with dimensionality of $height \times width \times time $; an image ensemble measured under multiple conditions can be represented by a higher order tensor with dimensionality of $ pixel \times person \times pose \times illumination$. Tensor factorization enables us to explicitly take into account the structure information by effectively capturing the multilinear interactions among multiple latent factors. Therefore, its theory and algorithms have been an active area of study during the past decade (see e.g.,~\cite{kolda2009tensor, Cichocki2009}), and have been successfully applied to various application fields, such as face recognition, social network analysis, image and video completion, and brain signal processing.  The two most popular tensor factorization frameworks are Tucker~\cite{tucker1966some} and CANDECOMP/PARAFAC (CP), also known as canonical polyadic decomposition (CPD)~\cite{Harshman1970foundations, bro1997parafac, sorensen2012canonical}.

Most existing tensor factorization methods assume that the tensor is complete, whereas the problem of missing data can arise in a variety of real-world applications. This issue has attracted a great deal of research interest in tensor completion in recent years. The objective of tensor factorization of incomplete data is to capture the underlying multilinear factors from only partially observed entries, which can in turn predict the missing entries. In~\cite{acar2011scalable}, CP factorization with missing data was formulated as a weighted least squares problem, termed CP weighted optimization (CPWOPT). A structured CPD using nonlinear least squares (CPNLS) was proposed in~\cite{sorber2013optimization}. In~\cite{kressner2013low}, geometric nonlinear conjugate gradient (geomCG) based on Riemannian optimization on the manifold of tensors were presented. However, as the number of missing entries increases, tensor factorization schemes tend to overfit the model because of an incorrectly specified tensor rank, resulting in severe deterioration of their predictive performance. In contrast, another popular technique is to exploit a low-rank assumption for recovering the missing entries; the rank minimization can be formulated as a convex optimization problem on a nuclear norm. This technique has been extended to higher order tensors by defining the nuclear norm of a tensor, yielding the tensor completion~\cite{liu2013tensorcompletion}. Some variants were also proposed, such as  a framework based on convex optimization and spectral regularization, which uses an inexact splitting method~\cite{signoretto2013learning}, and fast composite splitting algorithms (FCSA)~\cite{huang2011composite}. To improve the efficiency, the Douglas-Rachford splitting technique~\cite{gandy2011tensor}, nonlinear Gauss-Seidal method~\cite{tan2014tensor} were also investigated. Recently, the nuclear norm based optimization was also applied to a supervised tensor dimensionality reduction method~\cite{zhong2014large}. An alternative method for tensor completion is to employ adaptive sampling schemes~\cite{krishnamurthy2013low}.  However, the nuclear norm of a tensor is defined straightforwardly by a weighted sum of the nuclear norm of mode-$n$ matricizations, which is related to \emph{multilinear rank} rather than \emph{CP rank}. In addition, these completion-based methods cannot explicitly capture the underlying factors. Hence, a simultaneous tensor decomposition and completion (STDC) method was introduced in which a rank minimization technique was combined with Tucker decomposition~\cite{chen2013simul}. To improve completion accuracy, auxiliary information was also exploited in~\cite{narita2012tensor, chen2013simul}, which strongly depends on the specific application. It is also noteworthy that the rank minimization based on convex optimization of the nuclear norm is affected by tuning parameters, which may tend to over- or under-estimate the true tensor rank.

It is important to emphasize that our knowledge about the properties of \emph{CP rank}, defined by the minimum number of rank-one terms in CP decomposition, is surprisingly limited. There is no straightforward algorithm to compute the rank even for a given specific tensor, and the problem has been shown to be NP-complete~\cite{haastad1990tensor}. The lower and upper bound of tensor rank was studied in~\cite{alexeev2011tensor, burgisser2011geometric}. The ill-posedness of the best low-rank approximation of a tensor was investigated in~\cite{de2008tensor}. In fact, determining or even bounding the rank of an arbitrary tensor is quite difficult in contrast to the matrix rank~\cite{elizabeth2013tensor}, and this difficulty would be significantly exacerbated in the presence of missing data.

Probabilistic models for matrix/tensor factorization have attracted much interest in collaborative filtering and matrix/tensor completion. Probabilistic matrix factorization was proposed in~\cite{salakhutdinov2008probabilistic}, and its fully Bayesian treatment using Markow chain Monte Carlo (MCMC) inference was shown in~\cite{salakhutdinov2008bayesian} and using variational Bayesian inference in~\cite{babacan2012sparse,lim2007variational}. Further extensions of nonparametric and robust variants were presented in~\cite{lawrence2009non,lakshminarayanan2011robust}. The probabilistic frameworks of tensor factorization were presented in~\cite{chu2009probabilistic, gao2012probabilistic, xiong2010temporal}. Other variants include extensions of the exponential family model~\cite{hayashi2010exponential} and the nonparametric Bayesian model~\cite{qiinfinite}. However, the tensor rank or model complexity are often given by a tuning parameter selected by either maximum likelihood or cross-validations, which are computationally expensive and inaccurate. Another important issue is that the inference of factor matrices is performed by either point estimation, which is prone to overfitting, or MCMC inference, which tends to converge very slowly.

To address these issues, we propose a fully Bayesian probabilistic tensor factorization model according to the CP factorization framework. Our objective is to infer the underlying multilinear factors from a noisy incomplete tensor and the predictive distribution of missing entries, while the rank of the true latent tensor can be  determined automatically and implicitly. To achieve this, we specify a sparsity-inducing hierarchical prior over multiple factor matrices with individual hyperparameters associated to each latent dimension, such that the number of components in factor matrices is constrained to be minimum. All the model parameters, including noise precision, are considered to be latent variables over which the corresponding priors are placed. Due to complex interactions among multiple factors and fully Bayesian treatment, learning the model is analytically intractable. Thus, we resort to the variational Bayesian inference and derive a deterministic solution to approximate the posteriors of all the model parameters and hyperparameters. Our method is characterized as a tuning parameter-free approach that can effectively avoid parameter selections. The extensive experiments and comparisons on synthetic data illustrate the advantages of our approach in terms of rank determination, predictive capability, and robustness to overfitting. Moreover, several real-word applications, including image completion, restoration, and synthesis, demonstrate that our method outperforms state-of-the-art approaches, including both tensor factorization and tensor completion, in terms of the predictive performance.

The rest of this paper is organized as follows. In Section~\ref{sec:preliminaries}, preliminary multilinear operations and notations are presented. In Section~\ref{sec:BTF}, we introduce the probabilistic CP model specification and the model learning via Bayesian inference. A variant of our method using mixture priors is proposed in Section~\ref{sec:BCPF-MP}. In Section~\ref{sec:results}, we present the comprehensive experimental results for both synthetic data and real-world applications, followed by our conclusion in Section~\ref{sec:conclusions}.

\section{Preliminaries and Notations}
\label{sec:preliminaries}
The order of a tensor is the number of dimensions, also known as ways or modes. Vectors (first-order tensors) are denoted by boldface lowercase letters, e.g., $\mathbf{a}$. Matrices (second-order tensors) are denoted by boldface capital letters, e.g., $\mathbf{A}$. Higher-order tensors (order $\geq 3$) are denoted by boldface calligraphic letters, e.g., $\tensor{A}$. Given an $N$th-order tensor $\tensor{X}\in\mathbb{R}^{I_{1}\times I_{2}\times\cdots\times I_{N}}$, its $(i_1, i_2, \ldots, i_N)$th entry is denoted by $\mathcal{X}_{i_{1}i_{2}\ldots i_{N}}$, where the indices typically range from $1$ to their capital version, e.g., $i_{n}=1, 2,\ldots, I_{n}, \forall n\in[1,N]$.

The \emph{inner product of two tensors} is defined by $\langle \tensor{A}, \tensor{B}\rangle = \sum_{i_1,i_2,...,i_N} \mathcal{A}_{i_1i_2...i_N}\mathcal{B}_{i_1i_2...i_N}$, and the squared Frobenius norm by $\|\tensor{A}\|_F^2 = \langle \tensor{A}, \tensor{A}\rangle$. As an extension to $N$ variables, the \emph{generalized inner product} of a set of vectors, matrices, or tensors is defined as a sum of element-wise products. For example, given $\{\mathbf{A}^{(n)}|n=1,\ldots, N\}$, we define
\begin{equation}\label{eq:innerprodofmatrix}
\left\langle \mathbf{A}^{(1)}, \cdots, \mathbf{A}^{(N)} \right\rangle =\sum_{i,j} \prod_n A_{ij}^{(n)}.
\end{equation}

The \emph{Hadamard product} is an entrywise product of two vectors, matrices, or tensors of the same sizes. For instance, given two matrices, $\mathbf{A}\in\mathbb{R}^{I\times J}$ and $\mathbf{B}\in\mathbb{R}^{I\times J}$, their Hadamard product is a matrix of size $I\times J$ and is denoted by $\mathbf{A}\circledast\mathbf{B}$. Without loss of generality, the Hadamard product of a set of matrices can be simply denoted by
\begin{equation}
\hadamard_n \mathbf{A}^{(n)} = \mathbf{A}^{(1)}\circledast\mathbf{A}^{(2)}\circledast\cdots\circledast\mathbf{A}^{(N)}.
\end{equation}
The \emph{Kronecker product}\cite{kolda2009tensor} of matrices $\mathbf{A}\in\mathbb{R}^{I\times J}$ and $\mathbf{B}\in\mathbb{R}^{K\times L}$ is a matrix of size $IK \times JL$, denoted by $\mathbf{A}\otimes \mathbf{B}$. The \emph{Khatri-Rao product} of matrices $\mathbf{A}\in\mathbb{R}^{I\times K}$ and $\mathbf{B}\in\mathbb{R}^{J\times K}$ is a matrix of size $IJ\times K$, defined by a columnwise Kronecker product and denoted by $\mathbf{A}\odot\mathbf{B}$. In particular, the Khatri-Rao product of a set of matrices in reverse order is defined by
\begin{equation}
\bigodot_{n}\mathbf{A}^{(n)} = \mathbf{A}^{(N)}\odot\mathbf{A}^{(N-1)}\odot\cdots\odot\mathbf{A}^{(1)},
\end{equation}
while the Khatri-Rao product of a set of matrices, except the $n$th matrix, denoted by $\mathbf{A}^{(\backslash n)}$,  is
\begin{equation}
\small
\bigodot_{k\neq n}\mathbf{A}^{(k)} = \mathbf{A}^{(N)}\odot\cdots\odot\mathbf{A}^{(n+1)}\odot\mathbf{A}^{(n-1)}\odot\cdots\odot \mathbf{A}^{(1)}.
\end{equation}

\section{Bayesian Tensor Factorization}
\label{sec:BTF}
\subsection{Probabilistic Model and Priors}

Let $\tensor{Y}$ be an incomplete $N$th-order tensor of size $I_1\times I_2\times \cdots \times I_N$ with missing entries. The element $\mathcal{Y}_{i_1 i_2 \ldots i_N}$ is observed if $({i_1, i_2,\cdots, i_N}) \in \Omega$, where $\Omega$ denotes a set of indices.  For simplicity, we also define a binary tensor $\tensor{O}$ of the same size as $\tensor{Y}$ as an indicator of observed entries. We assume $\tensor{Y}$ is a noisy observation of true latent tensor $\tensor{X}$, that is, $\tensor{Y} = \tensor{X} + \tensor{\varepsilon}$, where the noise term is assumed to be an i.i.d. Gaussian distribution, i.e., $\tensor{\varepsilon}\sim\prod_{i_1,\ldots,i_N}\mathcal{N}(0,\tau^{-1})$, and the latent tensor $\tensor{X}$ can be exactly represented by a CP model, given by
\begin{equation}\label{eq:CPmodel}
\tensor{X} = \sum_{r=1}^R \mathbf{a}^{(1)}_{r} \circ\cdots\circ \mathbf{a}^{(N)}_{r} = [\![ \mathbf{A}^{(1)},\ldots, \mathbf{A}^{(N)} ]\!],
\end{equation}
where $\circ$  denotes the outer product of vectors and $[\![\cdots]\!]$ is a shorthand notation, also termed as the Kruskal operator. CP factorization can be interpreted as a sum of $R$ rank-one tensors, while the smallest integer $R$ is defined as \emph{CP rank}~\cite{kolda2009tensor}. $\{\mathbf{A}^{(n)}\}_{n=1}^N$ are a set of factor matrices where \mbox{mode-$n$} factor matrix $\mathbf{A}^{(n)}\in\mathbb{R}^{I_n\times R}$ can be denoted by row-wise or column-wise vectors
\begin{equation}\nonumber
\small
\begin{split}
\mathbf{A}^{(n)} =\left[\mathbf{a}^{(n)}_{1},\ldots,\mathbf{a}^{(n)}_{i_n},\ldots,\mathbf{a}^{(n)}_{I_n}\right]^T  =\left[\mathbf{a}^{(n)}_{\cdot 1},\ldots,\mathbf{a}^{(n)}_{\cdot r},\ldots,\mathbf{a}^{(n)}_{\cdot R} \right].
\end{split}
\end{equation}

\begin{figure}
\centering
  \includegraphics[width=0.25\textwidth]{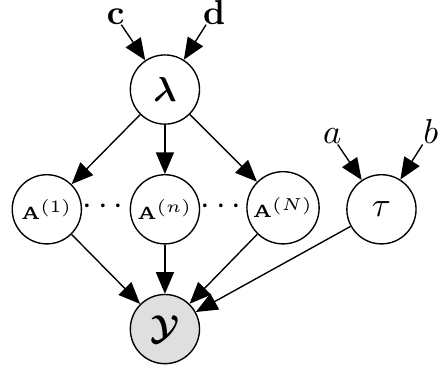}\\
  \caption{Probabilistic graphical model of Bayesian CP factorization of an $N$th-order tensor. }
  \label{fig:graphmodel}
\end{figure}

The CP generative model, together with noise assumption, directly give rise to the observation model, which is factorized over observed tensor elements
\begin{multline}\label{eq:observationModel}
p\Big(\tensor{Y}_{\Omega} \Big\vert \{\mathbf{A}^{(n)}\}_{n=1}^N, \tau\Big) =  \prod_{i_1=1}^{I_1}\cdots\prod_{i_N=1}^{I_N}\\
\mathcal{N}\left(\mathcal{Y}_{i_1 i_2 \ldots i_N} \middle\vert \left\langle\mathbf{a}^{(1)}_{i_1},\mathbf{a}^{(2)}_{i_2},\cdots,\mathbf{a}^{(N)}_{i_N}\right\rangle, \tau^{-1}\right)^{\mathcal{O}_{i_1\ldots i_N}},
\end{multline}
where the parameter $\tau$ denotes the noise precision, and $\left\langle\mathbf{a}^{(1)}_{i_1},\mathbf{a}^{(2)}_{i_2},\cdots,\mathbf{a}^{(N)}_{i_N}\right\rangle=\sum_r \prod_n {a}^{(n)}_{i_n r}$ denotes a generalized inner-product of $N$ vectors. The likelihood model in ($\ref{eq:observationModel}$) indicates that  $\mathcal{Y}_{i_1\cdots i_N}$ is generated by multiple $R$-dimensional latent vectors $\big\{\mathbf{a}^{(n)}_{i_n}\big\vert n=1,\ldots,N\big\}$, where each latent vector $\mathbf{a}_{i_n}^{(n)}$ contributes to a set of observations, i.e., a subtensor whose \mbox{mode-$n$} index is $i_n$. The essential difference between matrix  and  tensor factorization is that the inner product of $N\geq3$ vectors allows us to model the multilinear interaction structure, which however leads to many more difficulties in model learning.

In general, the effective dimensionality of the latent space, i.e., ${Rank_{CP}(\tensor{X})}=R$, is a tuning parameter whose selection is quite challenging and computational costly. Therefore, we seek an elegant  automatic model selection, which can not only infer the rank of the latent tensor $\tensor{X}$, but also effectively avoid overfitting. To achieve this, a set of continuous hyperparameters  are employed to control the variance related to each dimensionality of the latent space, respectively. Since the minimum $R$ is desired in the sense of low rank approximation, a sparsity-inducing prior is specified over these hyperparameters, resulting in it being possible to achieve automatic rank determination as a part of the Baybesian inference process. This technique is related to automatic relevance determination (ARD)~\cite{mackay1996bayesian} or sparse Bayesian learning~\cite{tipping2001sparse}. However, unlike the traditional methods that place the ARD prior over either latent variables or weight parameters, such as Bayesian principle component analysis~\cite{bishop1999bayesian}, our method considers all model parameters as latent variables over which a sparsity-inducing prior is placed with shared hyperparameters.

More specifically, we place a prior distribution over the latent factors, governed by  hyperparameters $\boldsymbol{\lambda} =[\lambda_1, \ldots, \lambda_R]$ where each $\lambda_r$ controls $r$th component in $\mathbf{A}^{(n)}$, which is
\begin{equation}\label{eq:priors}
p\big(\mathbf{A}^{(n)}\big\vert \boldsymbol\lambda \big) = \prod_{i_n=1}^{I_n} \mathcal{N}\big(\mathbf{a}_{i_n}^{(n)} \big\vert \mathbf{0}, \boldsymbol\Lambda^{-1} \big), \, \forall n\in [1,N],
\end{equation}
where $\boldsymbol\Lambda=\text{diag}(\boldsymbol\lambda)$ denotes the inverse covariance matrix, also known as the precision matrix, and is shared by latent factor matrices in all modes.
We can further define a hyperprior over $\boldsymbol\lambda$, which is factorized over latent dimensions
\begin{equation}\label{eq:hyperprior1}
p(\boldsymbol\lambda) = \prod_{r=1}^{R}\text{Ga}(\lambda_r \vert c^{r}_0, d^{r}_0),
\end{equation}
where $\text{Ga}(x\vert a,b) = \frac{b^a x^{a-1} e^{-bx}}{\Gamma(a)}$ denotes a Gamma distribution
 and $\Gamma(a)$ is the Gamma function.

Since the sparsity is enforced in the latent dimensions, the initialization point of the dimensionality of latent space (i.e., R) is usually set to its maximum possible value, while the effective dimensionality can be inferred automatically under a Bayesian inference framework. It should be noted that since the priors are shared across $N$ latent matrices, our framework can learn the same sparsity pattern for them, yielding the minimum number of rank-one terms. Therefore, our model can effectively infer the rank of tensor while performing the tensor factorization, which can be treated as a \emph{Bayesian low-rank tensor factorization}.

To complete the model with a fully Bayesian treatment, we also place a hyperprior over the noise precision $\tau$, that is,
\begin{equation}\label{eq:hyperprior2}
p(\tau) = \text{Ga}(\tau\vert a_0, b_0).
\end{equation}
For simplicity of notation, all unknowns including latent variables and \mbox{hyperparameters} are collected and denoted together by $\Theta =\{\mathbf{A}^{(1)},\ldots,\mathbf{A}^{(N)}, \boldsymbol\lambda, \tau \}$. The probabilistic graph model is illustrated in Fig.~\ref{fig:graphmodel}, from which we can easily write the joint distribution of the model as
\begin{equation}
\nonumber
\small
p(\tensor{Y}_\Omega, \Theta) = \condp{\tensor{Y}_\Omega} {\{\mathbf{A}^{(n)}\}_{n=1}^N, \tau} \prod_{n=1}^{N} \condp{\mathbf{A}^{(n)}}{\boldsymbol{\lambda}} p(\boldsymbol\lambda) p(\tau).
\end{equation}
By combining the likelihood in (\ref{eq:observationModel}), the priors of model parameters in (\ref{eq:priors}), and the hyperpriors in (\ref{eq:hyperprior1}) and (\ref{eq:hyperprior2}), the logarithm of the joint distribution is given by (see Sec.~1 of Appendix for details)
\begin{multline}\label{eq:logjointdistribution}
\ell(\Theta) = -\frac{\tau}{2} \left\|\tensor{O}\circledast\left(\tensor{Y}- [\![ \mathbf{A}^{(1)},\ldots, \mathbf{A}^{(N)} ]\!]\right)\right\|_F^2 \\
-\frac{1}{2} \text{Tr}\left(\boldsymbol\Lambda \sum_n \mathbf{A}^{(n)T}\mathbf{A}^{(n)}\right) + \left(\frac{M}{2} + a_0-1\right)\ln\tau \\
+\sum_r \left[\left(\frac{\sum_n I_n}{2} +(c_0^r-1) \right) \ln\lambda_r \right] \\
- \sum_r d_0^r\,\lambda_r  -b_0\tau + \text{const},
\end{multline}
where $M = \sum_{i_1,\ldots,i_N}\mathcal{O}_{i_1\ldots i_N}$ denotes the total number of observations. Without loss of generality, we can perform maximum a posteriori (MAP) estimation of $\Theta$ by maximizing (\ref{eq:logjointdistribution}), which is, to some extent, equivalent to optimizing a squared error function with regularizations imposed on the factor matrices and additional constraints imposed on the regularization parameters.

However, our objective is to develop a method that, in contrast to the point estimation, computes the full posterior distribution of all variables in $\Theta$ given the observed data, that is,
\begin{equation}\label{eq:posterior}
p(\Theta\vert\tensor{Y}_{\Omega})=  \frac{ p(\Theta,\tensor{Y}_\Omega)}{\int p(\Theta,\tensor{Y}_\Omega)\,d\Theta}.
\end{equation}
Based on the posterior distribution of $\Theta$, the predictive distribution over missing entries, denoted by $\tensor{Y}_{\backslash \Omega}$, can be inferred by
\begin{equation}\label{eq:predictiveform}
p(\tensor{Y}_{\backslash\Omega}\vert \tensor{Y}_{\Omega}) = \int  p(\tensor{Y}_{\backslash\Omega}\vert \Theta )p(\Theta\vert \tensor{Y}_\Omega)\, \text{d}\Theta.
\end{equation}

\subsection{Model Learning via Bayesian Inference}
An exact Bayesian inference in~(\ref{eq:posterior}) and (\ref{eq:predictiveform}) would integrate over all latent variables as well as hyperparameters, which is obviously analytically intractable. In this section, we describe the development of a deterministic approximate inference under variational Bayesian (VB) framework~\cite{winn2005variational, bishop2006pattern} to learn the probabilistic CP factorization model.

We therefore seek a distribution $q(\Theta)$ to approximate the true posterior distribution $p(\Theta\vert \tensor{Y}_\Omega)$ by minimizing  the KL divergence, that is,
\begin{multline}\label{eq:VBKL}
\text{KL}\big(q(\Theta)\big\vert\big\vert p(\Theta\vert \tensor{Y}_\Omega)\big) =\int q(\Theta) \ln \left\{\frac{q(\Theta)}{p(\Theta\vert \tensor{Y}_\Omega)} \right\} d\Theta\\
  =\ln p(\tensor{Y}_\Omega) - \int q(\Theta)\ln\left\{\frac{p(\tensor{Y}_\Omega,\Theta)}{q(\Theta)}\right\}d\Theta,
\end{multline}
where $\ln p(\tensor{Y}_\Omega)$ represents the model evidence, and its \emph{lower bound} is defined by $\mathcal{L}(q) = \int q(\Theta)\ln\left\{\frac{p(\tensor{Y}_\Omega,\Theta)}{q(\Theta)}\right\}d\Theta$. Since the model evidence is a constant, the maximum of the lower bound occurs when the KL divergence vanishes, which implies that $q(\Theta) =p(\Theta\vert \tensor{Y}_\Omega)$.

For the initial derivation, it will be assumed that the variational distribution is factorized w.r.t. each variable $\Theta_j$ and therefore can be written as
\begin{equation}
\label{eq:vbfactorization}
q(\Theta) = q_{\lambda}(\boldsymbol\lambda) q_{\tau}(\tau) \prod_{n=1}^{N} q_{n}\left(\mathbf{A}^{(n)}\right).
\end{equation}
It should be noted that this is the only assumption about the distribution, while the particular functional forms of the individual factors $q_j(\Theta_j)$ can be explicitly derived  in turn. The optimised form of the $j$th factor based on the maximization of $\mathcal{L}(q)$ is given by
\begin{equation}\label{eq:vbupdaterule}
\ln q_j(\Theta_j) = \mathbb{E}_{q_{(\Theta \backslash \Theta_j)}}[\ln p(\tensor{Y}_\Omega,\Theta)] + \text{const},
\end{equation}
where $\mathbb{E}_{q_{(\Theta \backslash \Theta_j)}}[\cdot]$ denotes an expectation w.r.t. the $q$ distributions over all variables except $\Theta_j$. Since the distributions of all parameters are drawn from the exponential family and are conjugate w.r.t. the distributions of their parents (see Fig.~\ref{fig:graphmodel}), we can derive the closed-form posterior update rules for each parameter in $\Theta$ by using (\ref{eq:vbupdaterule}).

\subsubsection{Posterior distribution of factor matrices}
As can be seen from the graphical model shown in Fig.~\ref{fig:graphmodel}, the inference of mode-$n$ factor matrix $\mathbf{A}^{(n)}$ can be performed by receiving the messages from observed data and its co-parents, including other factors $\mathbf{A}^{(k)}, k\neq n$ and the hyperparameter $\tau$, which are expressed by the likelihood term (\ref{eq:observationModel}), and incorporating the messages from its parents, which are expressed by the prior term (\ref{eq:priors}). By applying (\ref{eq:vbupdaterule}), it has been shown  that  their posteriors can be factorized as independent distributions of their rows, which are also Gaussian  (see Sec.~2 of Appendix  for details), given by
\begin{equation}\label{eq:qDistribution}
\begin{split}
q_n(\mathbf{A}^{(n)}) &= \prod_{i_n=1}^{I_n} \normalpdf{{\mathbf{a}}^{(n)}_{i_n} }{\tilde{\mathbf{a}}^{(n)}_{i_n} }{\mathbf{V}^{(n)}_{i_n} }, \, \forall n\in[1,N]
\end{split}
\end{equation}
where the posterior parameters can be updated by
\begin{equation}\label{eq:updatefactor}
\begin{split}
\tilde{\mathbf{a}}^{(n)}_{i_n} = \mathbb{E}_q[\tau] \mathbf{V}^{(n)}_{i_n} \mathbb{E}_q\big[ \mathbf{A}_{i_n}^{(\backslash n)T}\big] \text{vec}\left(\tensor{Y}_{\mathbb{I}(\mathcal{O}_{i_n}=1)}\right)\\
\mathbf{V}^{(n)}_{i_n} = \left(\mathbb{E}_q[\tau]  \mathbb{E}_q\big[\mathbf{A}_{i_n}^{(\backslash n)T}\mathbf{A}_{i_n}^{(\backslash n)}\big] + \mathbb{E}_q[\boldsymbol\Lambda] \right)^{-1},
\end{split}
\end{equation}
where $\tensor{Y}_{\mathbb{I}(\mathcal{O}_{i_n}=1)}$ is a sample function denoting a subset of the observed entries $\tensor{Y}_\Omega$, whose mode-$n$ index is $i_n$, i.e., the observed entries associated to the latent factor $\mathbf{a}_{i_n}^{(n)}$. The most complex term in (\ref{eq:updatefactor}) is related to
\begin{equation}
\mathbf{A}_{i_n}^{(\backslash n)T} = \Big(\bigodot_{k\neq n} \mathbf{A}^{(k)}\Big)^T_{\mathbb{I}(\mathcal{O}_{i_n}=1)},
\end{equation}
where $(\bigodot_{k\neq n} \mathbf{A}^{(k)})^T$ is of size $R\times\prod_{k\neq n} I_k $, and each column is computed by $\hadamard_{k\neq n}\mathbf{a}_{i_k}^{(k)}$ with varying mode-$k$ index $i_k$.
The symbol $(\cdot)_{\mathbb{I}(\mathcal{O}_{i_n}=1)}$ denotes a subset of columns sampled according to the subtensor $\text{vec}(\tensor{O}_{\cdots i_n\cdots}) =1$. Hence,
$\mathbb{E}_q[\mathbf{A}_{i_n}^{(\backslash n)T}\mathbf{A}_{i_n}^{(\backslash n)}]$ denotes the posterior covariance matrix of the Khatri-Rao product of latent factors in all modes except the $n$th-mode, and is computed by only the columns corresponding to the observed entries whose mode-$n$ index is $i_n$. In order to evaluate this posterior covariance matrix, first we need to introduce the following results.

\begin{theorem}\label{theorem:1}
Given a set of independent random matrices $\{\mathbf{A}^{(n)}\in\mathbb{R}^{I_n\times R}|n=1,\ldots,N\}$, we assume that $\forall n, \forall i_n$, the row vectors $\{\mathbf{a}^{(n)}_{i_n}\}$ are independent, then
\begin{equation}\label{eq:expectationofcovariance}
\small
\mathbb{E}\left[ \Big(\bigodot_n \mathbf{A}^{(n)}\Big)^T \Big(\bigodot_n \mathbf{A}^{(n)}\Big)\right] =\sum_{i_1,\ldots,i_N} \hadamard_n \left(\mathbb{E}\left[ \mathbf{a}^{(n)}_{i_n}\mathbf{a}^{(n)T}_{i_n}\right] \right)
\end{equation}
where $\mathbb{E}\left[ \mathbf{a}^{(n)}_{i_n}\mathbf{a}^{(n)T}_{i_n}\right] = \mathbb{E}[\mathbf{a}_{i_n}^{(n)}]\mathbb{E}[\mathbf{a}_{i_n}^{(n)T}] + \text{Var}\big(\mathbf{a}^{(n)}_{i_n}\big)$.
\begin{proof}
See Sec.~3 of Appendix  for details.
\end{proof}
\end{theorem}

For simplicity, we attempt to compute (\ref{eq:expectationofcovariance}) by multilinear operations. Let $\forall n$, $\mathbf{B}^{(n)}$ of size $I_n\times R^2$ denote an expectation of a quadratic form related to $\mathbf{A}^{(n)}$ by defining the
$i_n$th-row vector as
\begin{equation}\label{eq:DefineB}
\mathbf{b}^{(n)}_{i_n} = \text{vec}\left(\mathbb{E}_q\left[ \mathbf{a}^{(n)}_{i_n}\mathbf{a}^{(n)T}_{i_n}\right]\right) = \text{vec}\left(\tilde{\mathbf{a}}^{(n)}_{i_n} \tilde{\mathbf{a}}^{(n)T}_{i_n} + \mathbf{V}_{i_n}^{(n)}\right),
\end{equation}
then we have
\begin{equation}\label{eq:fastcomputation}
\begin{split}
\text{vec}\left(\sum_{i_1,\ldots,i_N} \hadamard_n \left(\mathbb{E}\left[ \mathbf{a}^{(n)}_{i_n}\mathbf{a}^{(n)T}_{i_n}\right] \right)\right) = \Big(\bigodot_{n}\mathbf{B}^{(n)}\Big)^T \; \mathbf{1}_{\prod_n {I_n}},
\end{split}
\end{equation}
where $\mathbf{1}_{\prod_n {I_n}}$ denotes a vector of length $\prod_n I_n$ and all elements are equal to one.

According to Theorem \ref{theorem:1} and the computation form in (\ref{eq:fastcomputation}), the term $\mathbb{E}_q\big[\mathbf{A}_{i_n}^{(\backslash n)T}\mathbf{A}_{i_n}^{(\backslash n)}\big]$ in (\ref{eq:updatefactor}) can be evaluated efficiently by
\begin{equation}\label{eq:ATA}
\begin{split}
\text{vec}\left(\mathbb{E}_q\big[\mathbf{A}_{i_n}^{(\backslash n)T}\mathbf{A}_{i_n}^{(\backslash n)}\big]\right)
= \Big(\bigodot_{k\neq n}\mathbf{B}^{(k)}\Big)^T \; \text{vec}(\tensor{O}_{\cdots i_n\cdots}),
\end{split}
\end{equation}
where  $\tensor{O}_{\cdots i_n\cdots}$ denotes a subtensor by fixing \mbox{model-$n$} index to $i_n$. It should be noted that the Khatri-Rao product is computed by all mode factors except the $n$th mode, while the sum is performed according to the indices of observations, implying that only factors that interact with $\mathbf{a}^{(n)}_{i_n}$ are taken into account. Another complicated part in (\ref{eq:updatefactor}) can  also be simplified by multilinear operations, i.e.,
\begin{align}\label{eq:ComputeMean}
\mathbb{E}_q&\big[ \mathbf{A}_{i_n}^{(\backslash n)T}\big] \text{vec}\left(\tensor{Y}_{\mathbb{I}(\mathcal{O}_{i_n}=1)}\right) \nonumber \\
&=\Big(\bigodot_{k\neq n} \mathbb{E}_q[{\mathbf{A}}^{(k)}] \Big)^T \text{vec}\big\{(\tensor{O}\circledast\tensor{Y})_{\cdots i_n\cdots}\big\}.
\end{align}

Finally, the variational posterior approximation of factor matrices can be updated by (\ref{eq:updatefactor}). On the basis of the approximated posterior, the posterior moments, including $\forall n, \forall i_n$, $\mathbb{E}_q\big[\mathbf{a}_{i_n}^{(n)}\big]$, $\text{Var}\big(\mathbf{a}_{i_n}^{(n)}\big)$, $\mathbb{E}_q\big[\mathbf{A}^{(n)}\big]$, and $\mathbb{E}_q\big[\mathbf{a}_{i_n}^{(n)}\mathbf{a}_{i_n}^{(n)T}\big], \mathbb{E}_q\big[\mathbf{a}_{i_n}^{(n)T}\mathbf{a}_{i_n}^{(n)}\big]$, can be easily evaluated, which are required by the inference of other hyperparameters in $\Theta$.

An intuitive interpretation of (\ref{eq:updatefactor}) is given as follows. The posterior covariance $\mathbf{V}^{(n)}_{i_n}$ is updated by combining the prior information $\mathbb{E}_q[\boldsymbol\Lambda]$ and the posterior information from other factor matrices computed by (\ref{eq:ATA}), while the tradeoff between these two terms is controlled by $\mathbb{E}_q[\tau]$ that is related to the quality of model fitting. In other words, the better fitness of the current model leads to more information from other factors than from prior information.
The posterior mean $\tilde{\mathbf{a}}^{(n)}_{i_n}$ is updated firstly by linear combination of all other factors, expressed by (\ref{eq:ComputeMean}), where the coefficients are observed values. This implies that the larger observation leads to more similarity of its corresponding latent factors. Then, $\tilde{\mathbf{a}}^{(n)}_{i_n}$ is rotated by $\mathbf{V}^{(n)}_{i_n}$ to obtain the property of sparsity and is scaled according to the model fitness $\mathbb{E}_q[\tau]$.

\subsubsection{Posterior distribution of hyperparameters $\boldsymbol\lambda$}
It should be noted that, instead of point estimation via optimizations, learning the posterior of $\boldsymbol\lambda$ is crucial for automatic rank determination. As seen in  Fig.~\ref{fig:graphmodel}, the inference of $\boldsymbol\lambda$ can be performed by receiving messages from $N$ factor matrices and incorporating the messages from its hyperprior. By applying (\ref{eq:vbupdaterule}), we can identify the posteriors of $\lambda_r,\forall r\in[1,R]$ as an independent Gamma distribution (see Sec.~4 of Appendix  for details),
\begin{equation}
q_{\boldsymbol\lambda}(\boldsymbol\lambda) = \prod_{r=1}^{R} \text{Ga}(\lambda_r \vert {c}_M^r, {d}_M^r ),
\end{equation}
where $c_M^r$, $d_M^r$ denote the posterior parameters learned from $M$ observations and can be updated by
\begin{equation}\label{eq:lambda2}
\begin{split}
c_M^r &= c_0^r + \frac{1}{2}\sum_{n=1}^{N} I_n,\\
d_M^r &= d_0^r + \frac{1}{2}\sum_{n=1}^{N} \mathbb{E}_q\left[\mathbf{a}^{(n)T}_{\cdot r}\mathbf{a}^{(n)}_{\cdot r}\right].
\end{split}
\end{equation}
The expectation  of the inner product of the $r$th component in mode-$n$ matrix w.r.t. $q$ distribution can be evaluated using the posterior parameters in (\ref{eq:qDistribution}), i.e.,
\begin{equation}\label{eq:lambda3}
\mathbb{E}_q\left[\mathbf{a}^{(n)T}_{\cdot r}\mathbf{a}^{(n)}_{\cdot r}\right] = \tilde{\mathbf{a}}^{(n)T}_{\cdot r} \tilde{\mathbf{a}}^{(n)}_{\cdot r}  + \sum_{i_n} \left(\mathbf{V}_{i_n}^{(n)}\right)_{rr}.
\end{equation}
By combining (\ref{eq:lambda2}) and (\ref{eq:lambda3}), we can further simplify the computation of $\mathbf{d}_M=[d_M^1,\ldots d_M^R]^T$ as
\begin{equation}
\mathbf{d}_M = \sum_{n=1}^{N} \left\{\text{diag}\left(\tilde{\mathbf{A}}^{(n)T}\tilde{\mathbf{A}}^{(n)} + \sum_{i_n} \mathbf{V}_{i_n}^{(n)} \right) \right\},
\end{equation}
where $\tilde{\mathbf{A}} = \mathbb{E}_q\big[\mathbf{A}^{(n)}\big]$. Hence, the posterior expectation can be obtained by $\mathbb{E}_q[\boldsymbol\lambda] = [c_M^1/d_M^1,\ldots,c_M^R/d_M^R]^T$, and thus, $\mathbb{E}_q[\boldsymbol\Lambda] = \text{diag}(\mathbb{E}_q[\boldsymbol\lambda])$.

An intuitive interpretation of (\ref{eq:lambda2}) is that $\lambda_r$ is updated by the sum of squared $L_2$-norm of $r$th component, expressed by (\ref{eq:lambda3}), from $N$ factor matrices. Therefore, the smaller of $\|\mathbf{a}_{\cdot r}\|_{2}^2$ leads to larger $\mathbb{E}_q[\lambda_r]$ and updated priors of factor matrices, which in turn enforces more strongly the $r$th component to be zero.

\subsubsection{Posterior distribution of hyperparameter $\tau$}
The inference of the noise precision $\tau$ can be performed by receiving the messages from observed data and its co-parents, including $N$ factor matrices, and incorporating the messages from its hyperprior. By applying (\ref{eq:vbupdaterule}), the variational posterior is a Gamma distribution (see Sec.~5 of Appendix  for details), given by
\begin{equation}
q_{\tau}(\tau) = \text{Ga}(\tau \vert {a}_M, b_M),
\end{equation}
where the posterior parameters can be updated by
\begin{equation}\label{eq:posteriorTau}
\begin{split}
{a}_M &= a_0+\frac{1}{2}\sum_{i_1,\ldots,i_N}\mathcal{O}_{i_1\ldots i_N}\\
{b}_M &= b_0+\frac{1}{2} \mathbb{E}_q\left[\left\|\tensor{O}\circledast\left(\tensor{Y}- [\![ \mathbf{A}^{(1)},\ldots, \mathbf{A}^{(N)} ]\!]\right)\right\|_F^2\right].
\end{split}
\end{equation}
However, the posterior expectation of model error in the above expression cannot be computed straightforwardly, and therefore, we need to introduce the following results.
\begin{theorem}\label{theorem:3}
 Assume a set of independent $R$-dimensional random vectors $\{\mathbf{x}^{(n)}|n=1,\ldots, N\}$, then
\begin{equation}
\small
\mathbb{E}\left[\left\langle\mathbf{x}^{(1)},\ldots,\mathbf{x}^{(N)}\right\rangle^2\right] = \left\langle \mathbb{E}\left[\mathbf{x}^{(1)} \mathbf{x}^{(1)T}\right], \ldots, \mathbb{E}\left[\mathbf{x}^{(N)}\mathbf{x}^{(N)T} \right]\right\rangle,
\end{equation}
where the left term denotes the expectation of the squared inner product of $N$ vectors, and the right term denotes the inner product of $N$ matrices, where each matrix of size $R\times R$ denotes an expectation of the outer product of the $n$th vector, respectively.
\begin{proof}
See Sec.~6  of Appendix for details.
\end{proof}
\end{theorem}

\begin{theorem}\label{theorem:4}
Given a set of independent random matrices $\{\mathbf{A}^{(n)}|n=1,\ldots,N\}$, we assume that $\forall n, \forall i_n$, the row vectors $\{\mathbf{a}_{i_n}^{(n)}\}$ are independent, then
\begin{multline}\label{eq:expX}
\mathbb{E}\left[ \left\| [\![ \mathbf{A}^{(1)},\ldots, \mathbf{A}^{(N)} ]\!]\right\|_F^2 \right]\\ = \sum_{i_1,\ldots,i_N} \left\langle \mathbb{E}\left[\mathbf{a}_{i_1}^{(1)}\mathbf{a}_{i_1}^{(1)T}\right],\ldots,\mathbb{E}\left[\mathbf{a}_{i_N}^{(N)}\mathbf{a}_{i_N}^{(N)T} \right]\right\rangle.
\end{multline}
Let $\mathbf{B}^{(n)}$ denote the expectation of a quadratic form related to $\mathbf{A}^{(n)}$ with $i_n$th-row vector $\mathbf{b}_{i_n}^{(n)} = \text{vec}\left(\mathbb{E}\left[\mathbf{a}_{i_n}^{(n)}\mathbf{a}_{i_n}^{(n)T}\right]\right)$; thus, (\ref{eq:expX}) can be computed by
\begin{multline}
\mathbb{E}\left[ \left\| [\![ \mathbf{A}^{(1)},\ldots, \mathbf{A}^{(N)} ]\!]\right\|_F^2 \right]=\mathbf{1}_{\prod_n I_n}^T \left(\bigodot_n \mathbf{B}^{(n)} \right)\mathbf{1}_{R^2}. \nonumber
\end{multline}
\begin{proof}
See Sec.~7 of Appendix  for details.
\end{proof}
\end{theorem}

From Theorems \ref{theorem:3} and \ref{theorem:4}, the posterior expectation term in (\ref{eq:posteriorTau}) can be evaluated explicitly. Due to the  missing entries in $\tensor{Y}$, the evaluation form is finally written as (see Sec.~8 of Appendix for details)
\begin{multline}\label{eq:modelerror}
\mathbb{E}_q\left[\left\|\tensor{O}\circledast\left(\tensor{Y}- [\![ \mathbf{A}^{(1)},\ldots, \mathbf{A}^{(N)} ]\!]\right)\right\|_F^2\right] \\
=\left\|\tensor{Y}_\Omega\right\|^2_F - 2\text{vec}^T(\tensor{Y}_\Omega)\text{vec}\left([\![ \tilde{\mathbf{A}}^{(1)},\ldots, \tilde{\mathbf{A}}^{(N)} ]\!]_\Omega\right) \\+ \text{vec}^T(\tensor{O}) \left(\bigodot_n \mathbf{B}^{(n)} \right)\mathbf{1}_{R^2},
\end{multline}
where $\tilde{\mathbf{A}}^{(n)}=\mathbb{E}_q\left[\mathbf{A}^{(n)}\right]$ and $\mathbf{B}^{(n)}$ is computed by (\ref{eq:DefineB}). Hence,
the posterior approximation of $\tau$  can be obtained by (\ref{eq:posteriorTau}) together with the posterior expectation $\mathbb{E}_q[\tau] = a_M/b_M$.

An intuitive interpretation of (\ref{eq:posteriorTau}) is straightforward. $a_M$ is related to the number of observations and $b_M$ is related to the residual of model fitting measured by the squared Frobenius norm on observed entries.

\subsubsection{Lower bound of model evidence}
The inference framework presented in the previous section can essentially maximize the lower bound of model evidence that is defined in (\ref{eq:VBKL}). Since the lower bound should not decrease at each iteration, it can be used to test for convergence. The lower bound of the log-marginal likelihood is computed by
\begin{equation}
\mathcal{L}(q) = \mathbb{E}_{q(\Theta)}[\ln p(\tensor{Y}_\Omega,\Theta)] + H(q(\Theta)),
\end{equation}
where the first term denotes the posterior expectation of joint distribution, and the second term denotes the entropy of posterior distributions.

Various terms in the lower bound are evaluated and derived by taking parametric forms of $q$ distribution, giving the following results (see Sec.~9 of Appendix  for details)
\begin{equation}\label{eq:lowerbound}
\begin{split}
&\mathcal{L}(q)=\\
&  -\frac{a_M}{2b_M}\mathbb{E}_q\left[\left\|\tensor{O}\circledast\left(\tensor{Y}- [\![ \mathbf{A}^{(1)},\ldots, \mathbf{A}^{(N)} ]\!]\right)\right\|_F^2\right] \\
&- \frac{1}{2} \text{Tr}\left\{ \tilde{\boldsymbol\Lambda} \sum_n\left(\tilde{\mathbf{A}}^{(n)T}\tilde{\mathbf{A}}^{(n)}+ \sum_{i_n} \mathbf{V}_{i_n}^{(n)}\right)\right\} \\
&+ \frac{1}{2}\sum_n\sum_{i_n} \left\{\ln\left|\mathbf{V}^{(n)}_{i_n}\right|\right\}+\sum_r \biggl\{  \ln \Gamma(c_M^r)\biggr\}\\
 &+ \sum_r\biggl\{c^r_M\left(1-\ln d_M^r -\frac{d^r_0}{d^r_M}\right) \bigg\}+\ln \Gamma(a_M)\\
 &+ a_M (1-\ln b_M -\frac{b_0}{b_M})  + \text{const}.
 \end{split}
\end{equation}
The posterior expectation of model error denoted by $\mathbb{E}_q[\cdot]$ can be computed using (\ref{eq:modelerror}).

An intuitive interpretation of (\ref{eq:lowerbound}) is as follows.  The first term is related to model residual; the second term is related to the weighted sum of squared $L_2$-norm of each component in factor matrices, while the uncertainty information is also considered;  the rest terms are related to  negative KL divergence between the posterior and prior distributions of hyperparameters.

\subsubsection{Initialization of model parameters}
The variational Bayesian inference is guaranteed to converge only to a local minimum. To avoid getting stuck in poor local solutions, it is important to choose an initialization point. In our model, the top level hyperparameters including $\mathbf{c}_0,\mathbf{d}_0$, $a_0,b_0$ are set to $10^{-6}$, resulting in a noninformative prior. Thus, we have $\mathbb{E}[\boldsymbol\Lambda]=\mathbf{I}$ and $\mathbb{E}[\tau]=1$.  For the factor matrices,   $\{\mathbb{E}[\mathbf{A}^{(n)}]\}_{n=1}^N$ can be initialized by two different strategies, one is randomly drawn from $\mathcal{N}(\mathbf{0},\mathbf{I})$ for $\mathbf{a}^{(n)}_{i_n}$, $ \forall i_n\in [1,I_n],\forall n\in [1,N]$. The other is set to $\mathbf{A}^{(n)}= \mathbf{U}^{(n)}\boldsymbol\Sigma^{(n)^{\frac{1}{2}}}$, where $\mathbf{U}^{(n)}$ denotes the left singular vectors and $\boldsymbol\Sigma^{(n)}$ denotes the diagonal singular values matrix, obtained by SVD of \mbox{mode-$n$} matricization of tensor $\tensor{Y}$. The covariance matrix $\mathbf{V}^{(n)}$ is simply set to $\mathbf{I}$. The tensor rank $R$ is usually initialized by the weak upper bound on its maximum rank, i.e., $R\leq \min_n P_n$, where $P_n= \prod_{i\neq n} I_i$. In practice, we can also manually define the initialization value of $R$  for computational efficiency.


\begin{algorithm}[tb]
   \caption{Fully Bayesian CP Factorization (FBCP) }
   \label{alg:FBCP}
\begin{algorithmic}
   \STATE {\bfseries Input:}  an $N$th-order incomplete tensor $\tensor{Y}_\Omega$ and an indicator tensor $\tensor{O}$.
   \STATE {\bfseries Initialization:} $\tilde{\mathbf{A}}^{(n)},\mathbf{V}_{i_n}^{(n)}, \forall i_n\in[1,I_n],\forall n\in[1,N]$,  $a_0,b_0,\mathbf{c}_0,\mathbf{d}_0$, and $\tau=a_0/b_0$, $\lambda_r=c_0^r/d_0^r, \forall r\in[1,R]$.
   \REPEAT
   \FOR{$n=1$ {\bfseries to} $N$}
   \STATE Update the posterior $q(\mathbf{A}^{(n)})$ using (\ref{eq:updatefactor});
   \ENDFOR
   \STATE Update the posterior  $q(\boldsymbol\lambda)$ using (\ref{eq:lambda2});
   \STATE Update the posterior  $q(\tau)$ using (\ref{eq:posteriorTau});
   \STATE Evaluate the lower bound using (\ref{eq:lowerbound});
   \STATE Reduce rank $R$ by eliminating zero-components of $\left\{\mathbf{A}^{(n)}\right\}$ (an optional procedure);
   \UNTIL{convergence.}
   \STATE Computation of  predictive distributions using (\ref{eq:predictive}).
\end{algorithmic}
\end{algorithm}

\subsubsection{Interpretaion of automatic rank determination}
The entire procedure of model inference is summarized in Algorithm~\ref{alg:FBCP}. It should be noted that tensor rank is determined automatically and implicitly. More specifically, updating $\boldsymbol\lambda$ in each iteration results in a new prior over $\{\mathbf{A}^{(n)}\}$, and then, $\{\mathbf{A}^{(n)}\}$ can be updated using this new prior in the subsequent iteration, which in turn affects $\boldsymbol\lambda$. Hence, if the posterior mean of $\lambda_r$ becomes very large, the $r$th components in $\{\mathbf{A}^{(n)}\},\forall n\in[1, N]$ are forced to be zero because of their prior information, and the tensor rank can be obtained by simply counting the number of non-zero components in the factor matrices. For implementation of the algorithm, we can keep the size of $\{\mathbf{A}^{(n)}\}$ unchanged during iterations; an alternative method is to eliminate the zero-components of $\{\mathbf{A}^{(n)}\}$ after each iteration.

\subsection{Predictive Distribution}
The predictive distributions over missing entries, given observed entries, can be approximated by using variational posterior distribution, that is,
\begin{equation}\label{eq:predictive}
\small
\begin{split}
&p(\mathcal{Y}_{i_1\ldots i_N} \vert \tensor{Y}_{\Omega}) = \int p(\mathcal{Y}_{i_1 \ldots i_N} \vert \Theta)p(\Theta\vert \tensor{Y}_{\Omega}) \, \text{d}\Theta\\
&\simeq\int\!\!\!\int p\left(\mathcal{Y}_{i_1\ldots i_N} \middle\vert \left\{\mathbf{a}^{(n)}_{i_n}\right\}, \tau^{-1}\right) q\left(\left\{\mathbf{a}^{(n)}_{i_n}\right\}\right) q(\tau) \,\text{d}\left\{\mathbf{a}^{(n)}_{i_n}\right\}\,\text{d}\tau.
\end{split}
\end{equation}

We can now approximate these integrations, yielding a Student's t-distribution $\mathcal{Y}_{i_1\ldots i_N} \vert \tensor{Y}_{\Omega}\sim \mathcal{T}(\tilde{\mathcal{Y}}_{i_1\ldots i_N}, \mathcal{S}_{i_1\ldots i_N}, \nu_y)$ (see Sec.~10 of Appendix for details)
with its parameters given by
\begin{equation}
\nonumber
\small
\begin{split}
\tilde{\mathcal{Y}}_{i_1\ldots i_N} &= \left\langle\tilde{\mathbf{a}}^{(1)}_{i_1},\cdots , \tilde{\mathbf{a}}^{(n)}_{i_N}\right\rangle, \quad \nu_y = 2a_M,\\
 \mathcal{S}_{i_1\ldots i_N} &= \left\{\frac{b_M}{a_M} + \sum_n \left\{ \left(\hadamard_{k\neq n}\tilde{\mathbf{a}}_{i_k}^{(k)}\right)^T \mathbf{V}^{(n)}_{i_n} \left(\hadamard_{k\neq n}\tilde{\mathbf{a}}_{i_k}^{(k)}\right)\right\}\right\}^{-1}.
\end{split}
\end{equation}
Thus, the predictive variance can be obtained by $\text{Var}(\mathcal{Y}_{i_1\ldots i_N}) = \frac{\nu_y}{\nu_y-2}\mathcal{S}_{i_1\ldots i_N}^{-1}$.

\subsection{Computational Complexity}
\label{sec:computation}
The computation cost of the $N$ factor matrices in (\ref{eq:updatefactor}) is $O(NR^2M + R^3 \sum_n I_n)$, where $N$ is the order of the  tensor, $M$ denotes the number of observations, i.e., the input data size. $R$ is the number of latent components in each $\mathbf{A}^{(n)}$, i.e., model complexity or tensor rank, and is generally much smaller than the data size, i.e., $R\ll M$. Hence, it has linear complexity w.r.t. the data size and polynomial complexity w.r.t. the model complexity. It should be noted that, because of the automatic model selection, the excessive latent components are pruned out in the first few iterations such that $R$ reduces rapidly in practice.  The computation cost of the hyperparameter $\boldsymbol\lambda$ in (\ref{eq:lambda2}) is $O(R^2\sum_n I_n)$, which is dominated by the model complexity, while the computation cost of noise precision $\tau$ in (\ref{eq:posteriorTau}) is $O(R^2 M)$. Therefore, the overall complexity of our algorithm is $O(NR^2M + R^3)$, which scales linearly with the data size but polynomially with the model complexity. 


\subsection{Discussion of Advantages}
\label{sec:discussions}
The advantages of our method are discussed as follows:
\begin{itemize}
  \item The \emph{automatic determination of \emph{CP} rank} enables us to obtain an optimal low-rank tensor approximation, even from a highly noisy and incomplete tensor.
  \item Our method is characterized as a \emph{tuning parameter-free} approach and all model parameters can be inferred   from the observed data, which avoids the computational expensive parameter selection procedure. In contrast, the existing tensor factorization methods require a predefined rank, while the tensor completion methods based on nuclear norm require several tuning parameters.
  \item The \emph{uncertainty information} over both latent factors and predictions of missing entries can be inferred by our method, while most existing tensor factorization and completion methods provide only the point estimations.
  \item An efficient and deterministic Bayesian inference is developed for model learning, which empirically shows a \emph{fast convergence}.
\end{itemize}

\section{Mixture Factor Priors}
\label{sec:BCPF-MP}
The low-rank assumption is powerful in general cases, however if the tensor data does not satisfy an intrinsic low-rank structure and a large amount of entries are missing, it may yield an oversimplified model. In this section, we present a variant of Bayesian CP factorization model which can take into account the local similarity in addition to the low-rank assumption. 

We specify a Gaussian mixture prior over factor matrices such that the prior distribution in (\ref{eq:priors}) can be rewritten as $\forall i_n\in[1, I_n],\forall n\in [1,N]$,
\begin{equation}\label{eq:MP}
\small
\nonumber
p\big(\mathbf{a}^{(n)}_{i_n}\big\vert \boldsymbol\lambda, \{\mathbf{a}^{(n)}_{k}\}\big) =  w_{i_n,i_n}\, \mathcal{N}\big( \mathbf{0}, \boldsymbol\Lambda^{-1} \big) + \sum_{k\neq i_n} w_{i_n,k} \, \mathcal{N}\big(\mathbf{a}_{k}^{(n)}, \beta_k^{-1} \mathbf{I}\big),
\end{equation}
where $ \sum_k w_{i_n,k} = 1$. This indicates that the $i_n$th row vector $a_{i_n}^{(n)}$ is similar to $k$th row vectors with the probability of $w_{i_n,k}$. Based on our assumption that the adjacent rows are highly correlated, we can define the mixture coefficients by $w_{i,j} = z_i \text{exp}(-|i-j|^2)$ where $z_i = 1/\sum_{j}\text{exp}(-|i-j|^2)$ is used to ensure the sum of mixture coefficients to be 1. For model learning, we can easily verify that the posterior distribution is also a mixture distribution. For simplicity, we set $\forall k, \beta_k =0$, thus the posterior mean of factor matrix can be updated firstly by (\ref{eq:updatefactor}) and then applying $\mathbb{E}_q[\mathbf{A}^{(n)}] \leftarrow \mathbf{W}\mathbb{E}_q[\mathbf{A}^{(n)}]$, while the posterior covariance $\{\mathbf{V}_{i_n}^{(n)}\}_{i_n=1}^{I_n}$ keep unchanged. Furthermore, the inference of all other variables do not need any changes.

\section{Experimental Results}
\label{sec:results}

We conducted extensive experiments using both synthetic data and real-world applications, and compared our fully Bayesian CP factorization (FBCP)\footnote{Matlab codes are available at \url{http://www.bsp.brain.riken.jp/~qibin/homepage/BayesTensorFactorization.html} } with seven state-of-the-art methods. Tensor factorization based scheme includes CPWOPT~\cite{acar2011scalable} and CPNLS~\cite{tensorlab2013,sorber2013optimization}, while the completion based scheme includes HaLRTC and FaLRTC~\cite{liu2013tensorcompletion}, FCSA~\cite{huang2011composite}, hard-completion (HardC.)~\cite{signoretto2013learning}, geomCG~\cite{kressner2013low} and STDC~\cite{chen2013simul}. Our objective when using synthetic data was to validate our method from several aspects: i) capability of rank determination; ii) reconstruction performance given a complete tensor; iii) predictive performance over missing entries given an incomplete tensor. Two real-world applications including image inpainting and facial image synthesis were used for demonstration. All experiments were performed by a PC (Intel Xeon(R) 3.3GHz, 64GB memory).

\subsection{Validation on Synthetic Data}
The synthetic tensor data is generated by the following procedure. $N$ factor matrices  $\{\mathbf{A}^{(n)}\}_{n=1}^N$  are drawn from a standard normal distribution, i.e., $\forall n, \forall i_n, \mathbf{a}^{(n)}_{i_n}\sim\mathcal{N}(\mathbf{0},\mathbf{I}_R)$. Then, the true tensor is constructed by $\tensor{X}=[\![\mathbf{A}^{1},\ldots,\mathbf{A}^{(N)}]\!]$, and an observed tensor by $\tensor{Y}=\tensor{X}+ \tensor{\varepsilon}$, where $\tensor{\varepsilon}\sim \prod_{i_1,\ldots,i_N}\mathcal{N}(0,\sigma^2_{\tensor{\varepsilon}})$ denotes i.i.d. additive noise. The missing entries, chosen uniformly, are marked by an indicator tensor $\tensor{O}$.

\begin{figure}
\centering
  \includegraphics[width=0.48\textwidth]{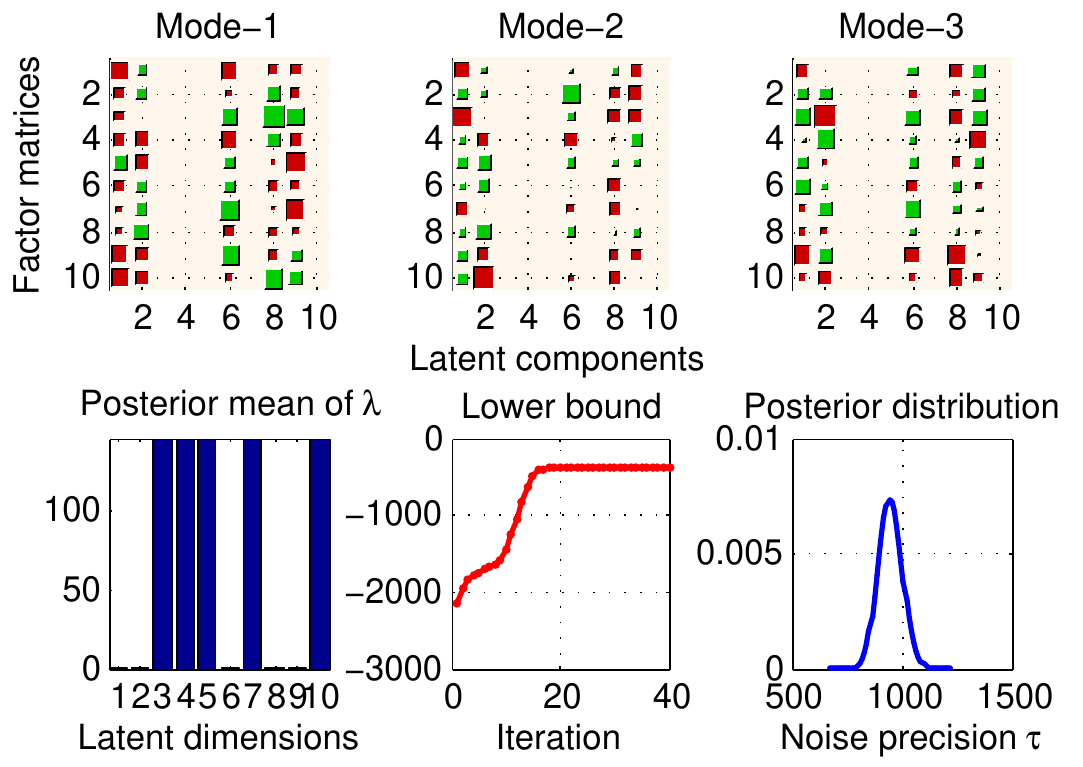}\\
  \caption{A toy example illustrating FBCP applied on an incomplete tensor. The top row shows Hinton diagram of factor matrices, where the color and size of each square represent the sign and magnitude of the value, respectively.  The bottom row shows the posterior of $\boldsymbol\lambda$, the lower bound of model evidence, and the posterior of $\tau$ from left to right. }
  \label{fig:demo}
\end{figure}

\subsubsection{A toy example}
To illustrate our model, we provide two demo videos in the supplemental materials. A true latent tensor $\tensor{X}$ is of size $10\times 10 \times 10$ with \emph{CP} rank $R=5$, the noise parameter was $\sigma^2_{\tensor{\varepsilon}}=0.001$, and $40\%$ of entries were missing. Then, we applied our method with the initial rank being set to 10. As shown in Fig.~\ref{fig:demo}, three factor matrices are inferred in which five components are effectively pruned out, resulting in correct estimation of tensor rank. The lower bound of model evidence increases monotonically, which indicates the effectiveness and convergence of our algorithm. Finally, the posterior of noise precision $\tau\approx 1000$ implies the method's capability of denoising and the estimation of $\sigma^2_{\tensor{\varepsilon}}\approx0.001$, $\text{SNR}= 10\log \frac{ \sigma^2_{\tensor{X}}} {\tau^{-1}} $.

\subsubsection{Automatic determination of tensor rank}

\begin{figure}[ht]
\centering
\subfigure[]{
   \includegraphics[width=0.46\columnwidth] {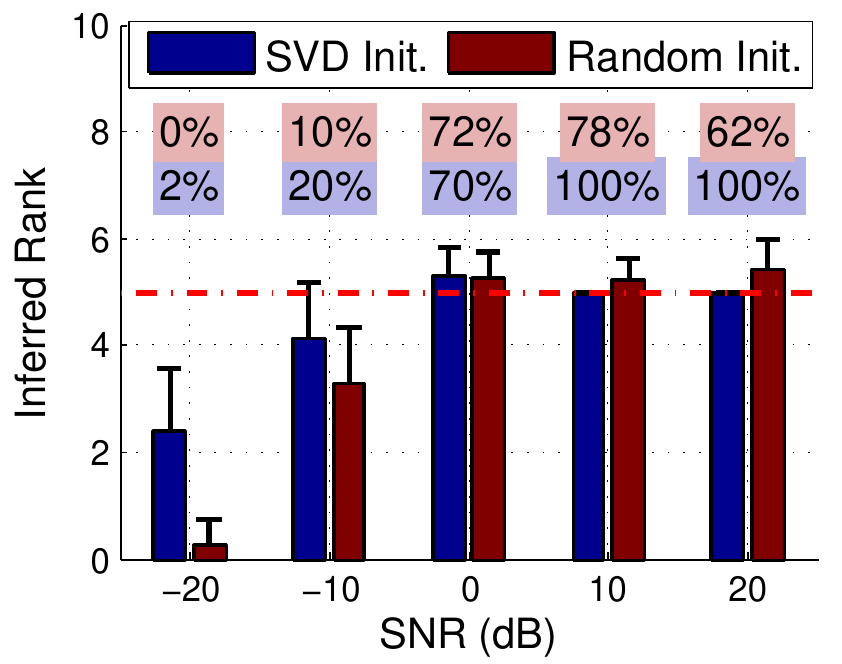}
   \label{fig:rank1}
 }
 \subfigure[]{
   \includegraphics[width=0.46\columnwidth] {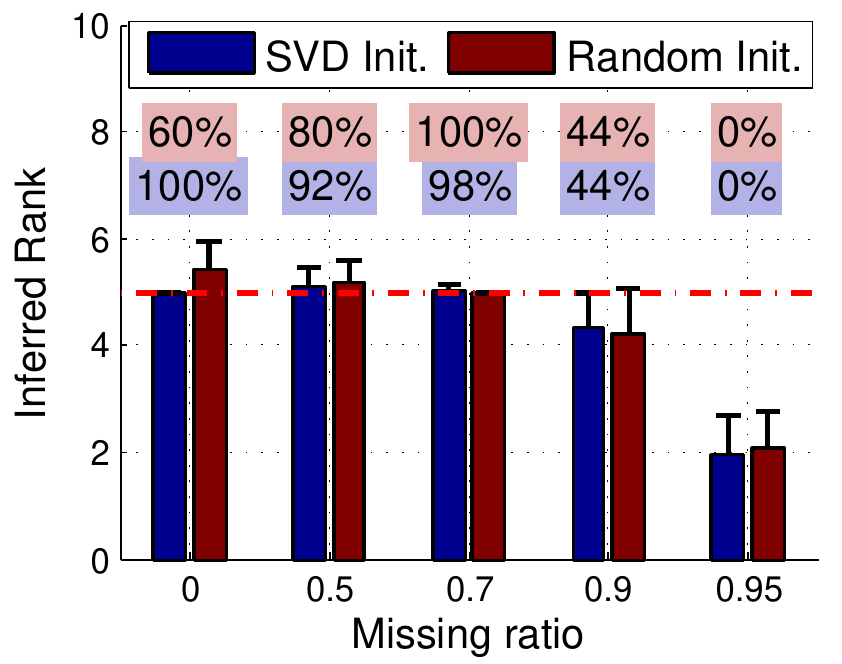}
   \label{fig:rank2}
   }
 \subfigure[]{
   \includegraphics[width=0.46\columnwidth] {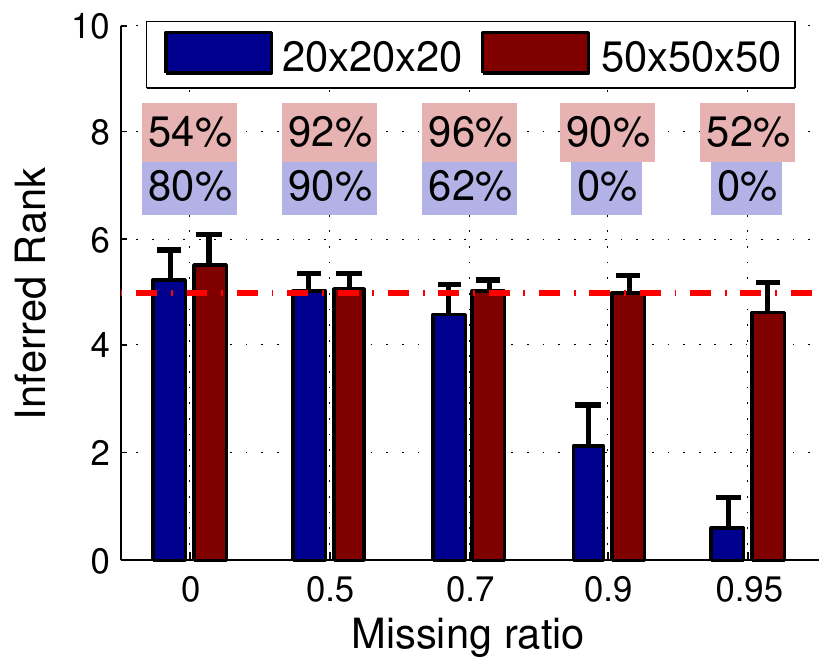}
   \label{fig:rank3}
   }
   \subfigure[]{
   \includegraphics[width=0.46\columnwidth] {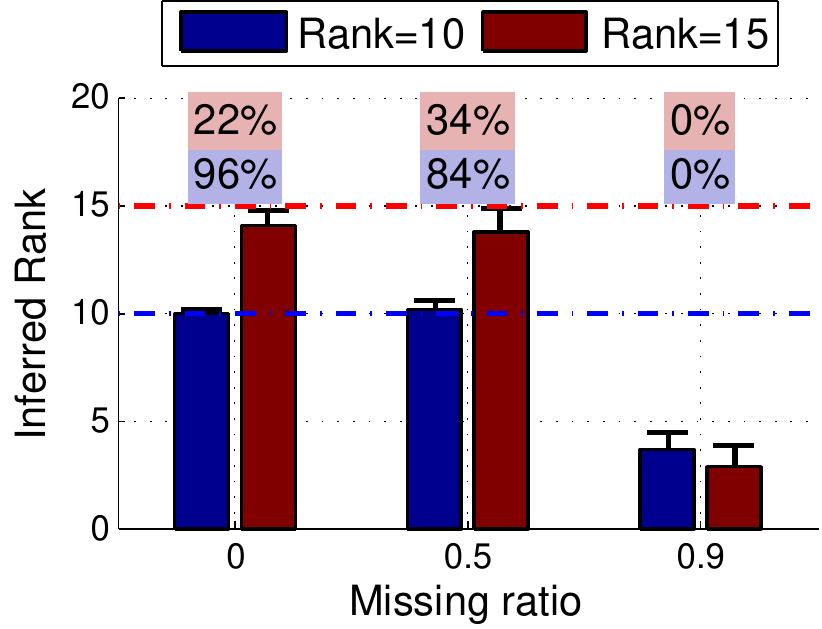}
   \label{fig:rank4}
 }
\caption{Determination of tensor rank  under varying experimental conditions. Each vertical bar shows the mean and standard deviation of estimations from 50 repetitions, while the accuracy of detections is shown on the top of the corresponding bar. The red and blue horizontal dash dotted lines indicate the true tensor rank.  }
\label{fig:tensorrank}
\end{figure}

To evaluate the automatic determination of tensor rank (i.e., \emph{CP rank}), extensive simulations were performed under varying experimental conditions related to tensor size, tensor rank, noise level, missing ratio, and the initialization method of factor matrices (e.g., SVD or random sample). Each result is evaluated by 50  repetitions  corresponding to 50 different tensors generated under the same criterion.  There are four groups of experiments. (A) Given complete tensors of size $20\times 20 \times 20$ with $R=5$, the evaluations were performed under varying noise levels and by two different initializations (see Fig.~\ref{fig:rank1}). (B) Given incomplete tensors of size $20\times 20\times 20$ with $R=5$ and SNR=20 dB, the evaluations were performed under five different missing ratios, and by different initializations (see Fig.~\ref{fig:rank2}). (C) Given incomplete tensors with $R=5$ and SNR=0 dB, the evaluations were performed under varying missing ratios and two different tensor sizes (see Fig.~\ref{fig:rank3}). (D) Given incomplete tensors of size $20\times 20\times 20$ with SNR=20 dB, the evaluations were performed under varying missing ratios and two different true ranks (see Fig.~\ref{fig:rank4}).

From Fig.~\ref{fig:tensorrank}, we observe that SVD initialization is slightly better than random initialization in terms of the determination of tensor rank. If the tensor is complete, our model can detect the true tensor rank with $100\%$ accuracy when SNR$\geq$10 dB. Although the accuracy decreased to 70\% under a high noise level of 0 dB, the error deviation is only $\pm 1$. On the other hand, if the tensor is incomplete and almost free of noise, the detection rate is 100\%, when missing ratio is 0.7, and is 44\% with an error deviation of only $\pm 1$, even under a high missing ratio of 0.9. As both missing data and high noise level are presented, our model can achieve 90\% accuracy under the condition of SNR=0 dB and 0.5 missing ratio. It should be noted that, when the data size is larger, such as $50\times 50\times 50$, our model can achieve 90\% accuracy, even when SNR=0 dB and the missing ratio is 0.9. If the true rank is larger, such as $R=15$, the model can correctly recover the rank from a complete tensor, but fails to do so when the missing ratio is larger than 0.5.

We can conclude from these results that the determination of the tensor rank depends primarily on the number of observed entries and the true tensor rank. In general, more observations are necessary if the tensor rank is larger; however, when high-level noise occurs, the excessive number of observations may not be helpful for rank determination.

\subsubsection{Predictive performance}
\begin{figure}[h]
\centering
  \includegraphics[width=0.42\textwidth]{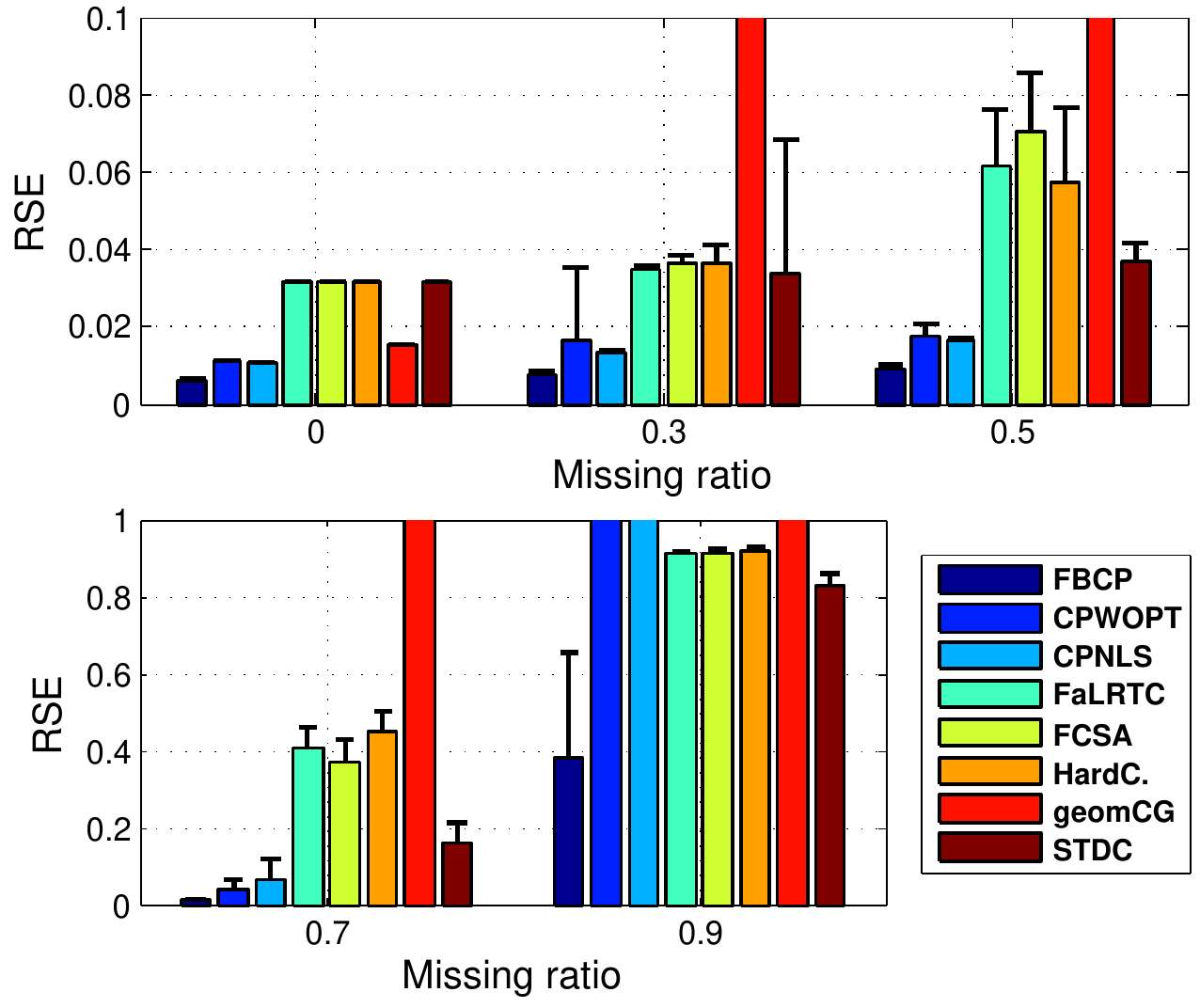}\\
  \caption{Predictive performance when SNR=30 dB.}
  \label{fig:RSEmissing30dB}
\end{figure}

In this experiment, we considered incomplete tensors of size $20\times 20\times 20$ generated by the true rank $R=5$ and SNR=30 dB under varying missing ratios. The initial rank was set to 10. The relative standard error $RSE = \frac{\|\hat{\tensor{X}}-\tensor{X}\|_F}{\|\tensor{X}\|_F}$, where $\hat{\tensor{X}}$ denotes the estimation of the true tensor $\tensor{X}$, was used to evaluate the performance.  To ensure statistically consistent results, the performance is evaluated by 50 repetitions for each condition.  As shown in Fig.~\ref{fig:RSEmissing30dB}, our method significantly outperforms other algorithms under all missing ratios. Factorization-based methods, including CPWOPT, and CPNLS show a better performance than completion-based methods when the missing ratio is relatively small, while they perform worse than completion methods when the missing ratio is large, e.g., 0.9. FaLRTC, FCSA, and HardC. achieve similar performances, because they are all based on nuclear norm optimization. geomCG achieves a performance comparable with that of CWOPT and CPNLS when data is complete, while it fails as the missing ratio becomes high. This is because geomCG requires a large number of observations and precisely defined rank. It should be noted that STDC outperforms all algorithms except FBCP as the missing ratio becomes extremely high. These results demonstrate that FBCP, as a tensor factorization method, can be also effective for tensor completion, even when an extremely sparse tensor is presented.

We also conducted two additional experiments. One is the reconstruction from a complete tensor, the other is the tensor completion when the noise level is high, i.e., SNR =0 dB. The results of these two experiments are presented in Appendix (see Sec. 11, 12).

\subsection{Image Inpainting}
\begin{figure}[h]
\centering
  \includegraphics[width=0.4\textwidth]{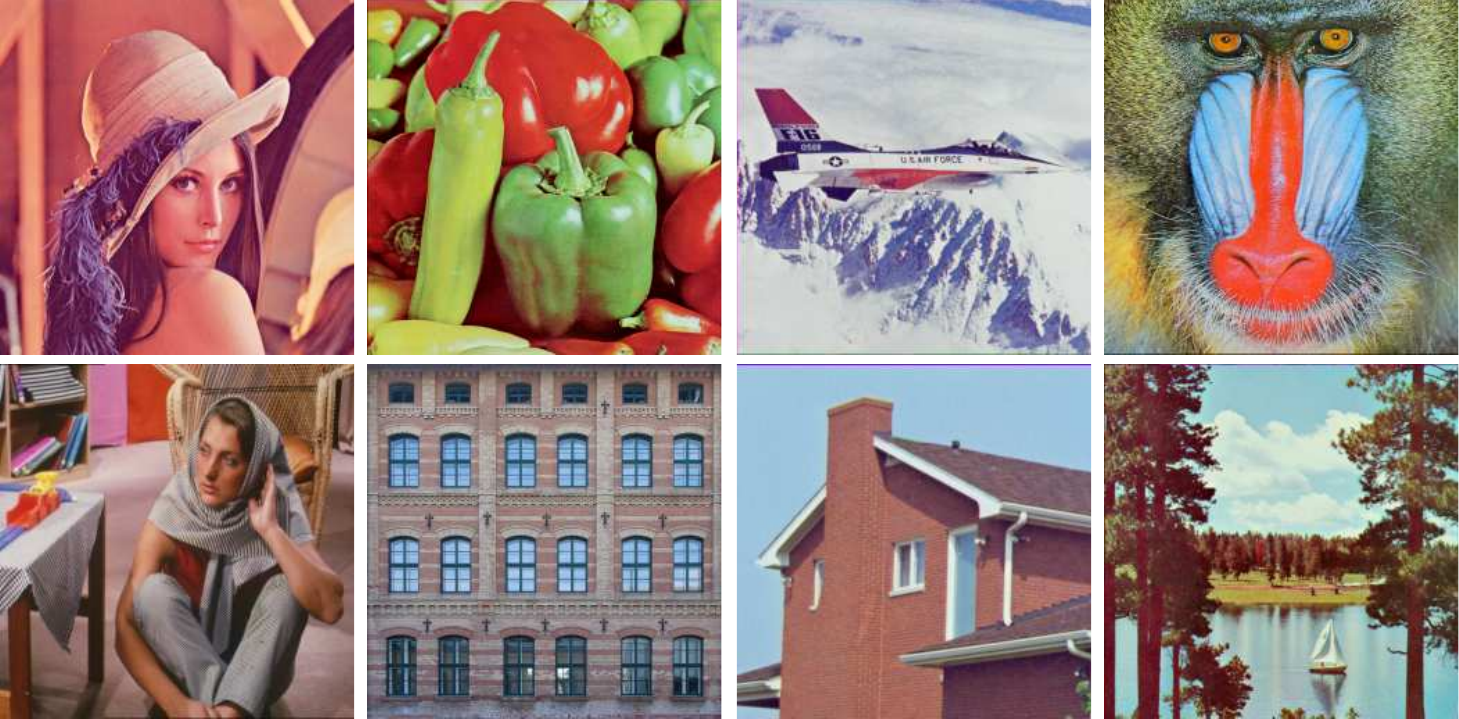}
  \caption{ Ground-truth of eight benchmark images.}
  \label{fig:GroundTruthImages}
\end{figure}

\begin{figure*}
\centering
  \includegraphics[width=0.95\textwidth]{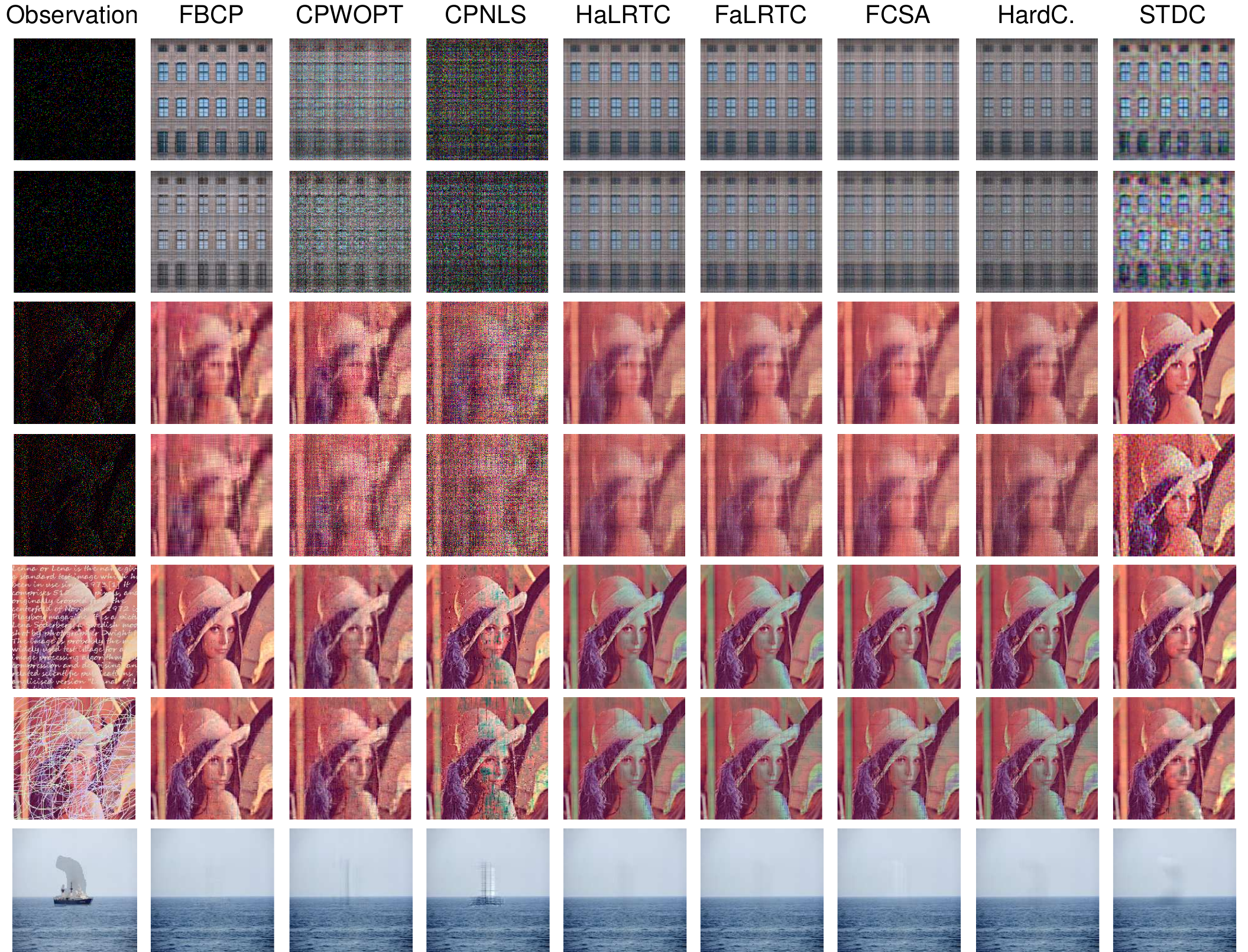}\\
  \caption{Visual effects of image inpainting. Seven examples shown from top to bottom are (1) facade image with 95\% missing; (2) facade image with 95\% missing and an additive noise; (3) lena image with 90\% missing; (4) lena image with 90\% missing and an additive noise; (5) lena image with superimposed text; (6) scribbled lena image; (7) an image of ocean with an object. }
  \label{fig:imageinpainting}
\end{figure*}

In this section,  the applications of image inpainting based on several benchmark images, shown in Fig.~\ref{fig:GroundTruthImages}, are used to evaluate and compare the performance of different methods. The colorful image can be represented by a third-order tensor of size $200\times 200\times 3$. We conducted various experiments under four groups of conditions. (A) \emph{Structural image with uniformly random missing pixels.} A building facade image with 95\% missing pixels under two noise conditions, i.e., noise free and SNR=5dB, were considered as observations. (B) \emph{Natural image with uniformly random missing pixels.} The Lena image of size $300\times 300$ with 90\% missing pixels under two noise conditions, i.e., noise free and SNR=10dB, were considered. (C) \emph{Non-random missing pixels.} We conducted two experiments for image restoration from a corrupted image: 1) The Lenna image corrupted by superimposed text was used as an observed image\footnote{A demo video is available in the supplemental materials.}. In practice, the location of text pixels are difficult to detect exactly; we can simply indicate missing entries by a value larger than 200 to ensure that the text pixels are completely missing. 2) The scrabbled Lenna image was used as an observed image and pixels with values larger than 200 can be marked as missing. (D) \emph{Object removal}. Given an image and a mask covering the object area, our goal was to complete the image without that object. The algorithm settings of compared methods are described as follows. For factorization-based methods, the initial rank was set to 50 in cases of (A) and (B) due to the high missing ratios, and 100 in cases of (C) and (D). For completion-based methods, the tuning parameters were chosen by multiple runs and performance evaluated on the ground-truth of missing pixels.


\begin{table}
\renewcommand{\arraystretch}{1.1}
\caption{\small Performance (RSEs) evaluated on missing pixels. ``NF'', ``N'' indicate noise free or noisy image. ``T'', ``S'' indicate the text corruption or scrabbled image. }
\label{tab:ImageCompletion}
\centering
{
\begin{tabular}{c c c  c c  c c }
\hline\hline
\multirow{2}{*}{Method} &  \multicolumn{2}{c}{Facade} & \multicolumn{2}{c}{Lenna} & \multicolumn{2}{c}{Non-random}   \\
\cmidrule(lr){2-3} \cmidrule(lr){4-5} \cmidrule(lr){6-7}
        & NF      & N      & NF      & N      & T      & S     \\
\hline
FBCP    &\bf 0.13    &\bf 0.17    & 0.17    &\bf 0.20    &\bf 0.13    &\bf 0.14   \\
CPWOPT  &0.33    &0.41    &0.25    &0.41    &0.18    &0.18   \\
CPNLS   &0.84    &0.84    &0.62    &0.73    &0.22    &0.30   \\
HaLRTC  &0.15    &0.21    &0.19    &0.21    &0.29    &0.28   \\
FaLRTC  &0.16    &0.21    &0.19    &0.22    &0.29    &0.29   \\
FCSA    &0.19    &0.21    &0.19    &0.21    &0.28    &0.28   \\
HardC.  &0.19    &0.25    &0.20    &0.23    &0.31    &0.30   \\
STDC    &0.14    &0.22    &\bf 0.11    &\bf 0.20    &0.15    &0.16   \\
\hline\hline
\end{tabular}
}
\vspace{-0.1in}
\end{table}

\begin{table*}[th]
\renewcommand{\arraystretch}{1.3}
\caption{The averaged recovery performance (RSE, PSNR, SSIM) and runtime (seconds) on eight images with missing rates of 70\%, 80\%, 90\% and 95\%. For methods that need to tune parameters, both the runtime with the best tuning parameter and the overall runtime are reported.}
\label{tab:averageperformance}
\centering
\resizebox{0.8\textwidth}{!}
{
\begin{tabular}{c | c | c | c | c |c |c |c |c |c }
 & & FBCP  & FBCP-MP & CPWOPT & STDC & HaLRTC & FaLRTC & FCSA & HardC.   \\
 \hline
 70\% & RSE &  0.1209 & \bf 0.0986 & 0.1493 & 0.1003 & 0.1205 & 0.1205 & 0.1406 & 0.1254  \\
      & PSNR & 25.13 & 26.78 & 23.35  & \bf 26.79 & 25.12   & 25.12 & 23.64   & 24.71\\
      & SSIM & 0.7546 & \bf 0.8531 & 0.6417 & 0.8245 & 0.7830 & 0.7831 & 0.7437 & 0.7730  \\
      & Runtime & 83  & 251    & 1807/2908  &  32/292    & 18/139    & 46/232    & \bf 9  &  36\\
\hline
 80\% & RSE & 0.1423  & \bf 0.1084 & 0.1700 & 0.1095 & 0.1479 & 0.1479 & 0.1675 & 0.1548 \\
      & PSNR & 23.09  & 25.36 & 21.52 & \bf 25.40  & 22.72  & 22.72 & 21.53 & 22.32 \\
      & SSIM & 0.6515  & \bf 0.7941 & 0.5567 & 0.7781 & 0.6716 & 0.6716  & 0.6410  & 0.6579  \\
      & Runtime &  76 &  196 & 590/2316 &  30/328 & 25/122 & 57/282 & \bf 9 & 39  \\
\hline
 90\% & RSE & 0.1878  & \bf 0.1295 & 0.2372 & 0.1316 & 0.1992 & 0.1995 & 0.2342 & 0.2121 \\
      & PSNR & 20.12  & \bf 23.26  & 18.08 & 23.21   & 19.62  & 19.61 & 18.09 & 19.11 \\
      & SSIM & 0.4842  & \bf 0.6956 & 0.3628 & 0.6950 & 0.5005 & 0.4998  & 0.4477  & 0.4790  \\
      & Runtime &  69 &  169 & 390/1475 &  32/378 & 21/127 & 61/307 & \bf 9 & 32  \\
\hline
 95\% & RSE & 0.2420  & \bf 0.1566 & 0.3231 & 0.1600 & 0.2549 & 0.2564 & 0.2777 & 0.2903 \\
      & PSNR & 17.76  & \bf 21.34 & 15.35 & 21.18 & 17.25  & 17.19 & 16.39 & 16.12\\
      & SSIM & 0.3455  & \bf 0.6031 & 0.2539 & 0.5810 & 0.3676 & 0.3649  & 0.3535  & 0.3369  \\
      & Runtime &  66 &  133 & 201/881 &  35/400 & 24/137 & 63/313 & \bf 8 & 32  \\
\end{tabular}
}
\end{table*}

The visual effects of image inpainting are shown in Fig.~\ref{fig:imageinpainting}, and the predictive performances are shown in Table~\ref{tab:ImageCompletion} where case (D) is not available due to the lack of ground-truth. In case (A), we observe that FBCP outperforms other methods for a structural image under an extremely high missing ratio and the superiority is more significant when an additive noise is involved. In case (B), observe that STDC obtains the best performance followed by FBCP that is better than other methods. However, STDC is severely degraded when noise is involved, and obtains the same performance as FBCP, while its visual quality is still much better than others. These indicate that the additional smooth constraints in STDC are suitable for natural image. In case (C), FBCP is superior to all other methods, followed by STDC. The completion-based methods obtain relatively smoother effects than factorization-based methods, but the global color of the image is not recovered well, resulting in a poor predictive performance. In case (D), FBCP obtains the most clean image by removing the object completely while the ghost effects appear in all other methods. HaLRTC, FaLRTC and FCSA outperform CPWOPT, CPNLS and STDC.

From these results we can conclude that the completion-based methods generally outperforms factorization-based methods for image completion. However, FBCP significantly improves ability of factorization-based scheme by automatic model selection and robustness to overfitting, resulting in potential applications for various image inpainting problems. The necessary number of observed entries mainly depends on the rank of true image. For instance, a structural image with an intrinsic low-rank need very fewer observations than a natural image. However only 10\% observed pixels from lena image are not sufficient to recover the whole image, which is caused by the low-rank assumption. This property is common for all these algorithms except STDC, because STDC employs an auxiliary information as additional constraints. The advantage of STDC has been shown for lena image, while its disadvantages are that the auxiliary information must be well designed for each specific application, which make it difficult to be applied to other types of data. In addition, STDC degrades in presence of the non-random missing pixels or noise. Moreover, the performance of HaLRTC and STDC are sensitive to tuning parameters that must be carefully selected for each specific condition. Therefore, a crucial drawback of completion-based scheme lies in the tuning parameters whose selection is quite challenging when the ground-truth of missing data is unknown.

Next, we perform image completion extensively on eight images in Fig.~\ref{fig:GroundTruthImages} with randomly missing pixels. Since most of these images are natural images on which the low-rank approximation cannot recover the missing pixels well, we apply the fully Bayesian CP with mixture priors (FBCP-MP) for comparison with FBCP and other related methods. For FBCP and FBCP-MP, the same initialization of $R=100$ was applied, while CPWOPT was performed by using the optimal ranks obtained from FBCP and FBCP-MP to show the best performance. The parameter selection for other methods was same with previous experiments. The size of all images is $256\times 256\times 3$. Table~\ref{tab:averageperformance} shows quantitative results in terms of recovery performance and runtime. Observe that FBCP-MP improves the performance of FBCP significantly and achieves the best recovery performance, especially in the case of high missing rate. The time costs of FBCP and FBCP-MP are comparable with completion-based methods and significantly lower than other tensor factorization method. STDC obtains the comparable performance with FBCP-MP, however the parameters must be manually tuned for the specific condition. More detailed results on each image are shown visually and quantitatively in the supplemental materials. These results demonstrate the effectiveness of mixture priors and advantages when the local similarity is taken into account in addition to the low-rank assumption.

\subsection{Facial Image Synthesis}
For recognition of face images captured from surveillance videos, the ideal solution is to create a robust classifier that is invariant to some factors, such as pose and illumination. Hence, there arises the question whether we can generate novel facial images under multiple conditions given images under other conditions.  Tensors are highly suitable for modeling a multifactor image ensemble, and therefore, we introduce a novel application of facial image synthesis that utilizes tensor factorization approaches.

\begin{figure}[h]
\centering
  \includegraphics[width=0.3\textwidth]{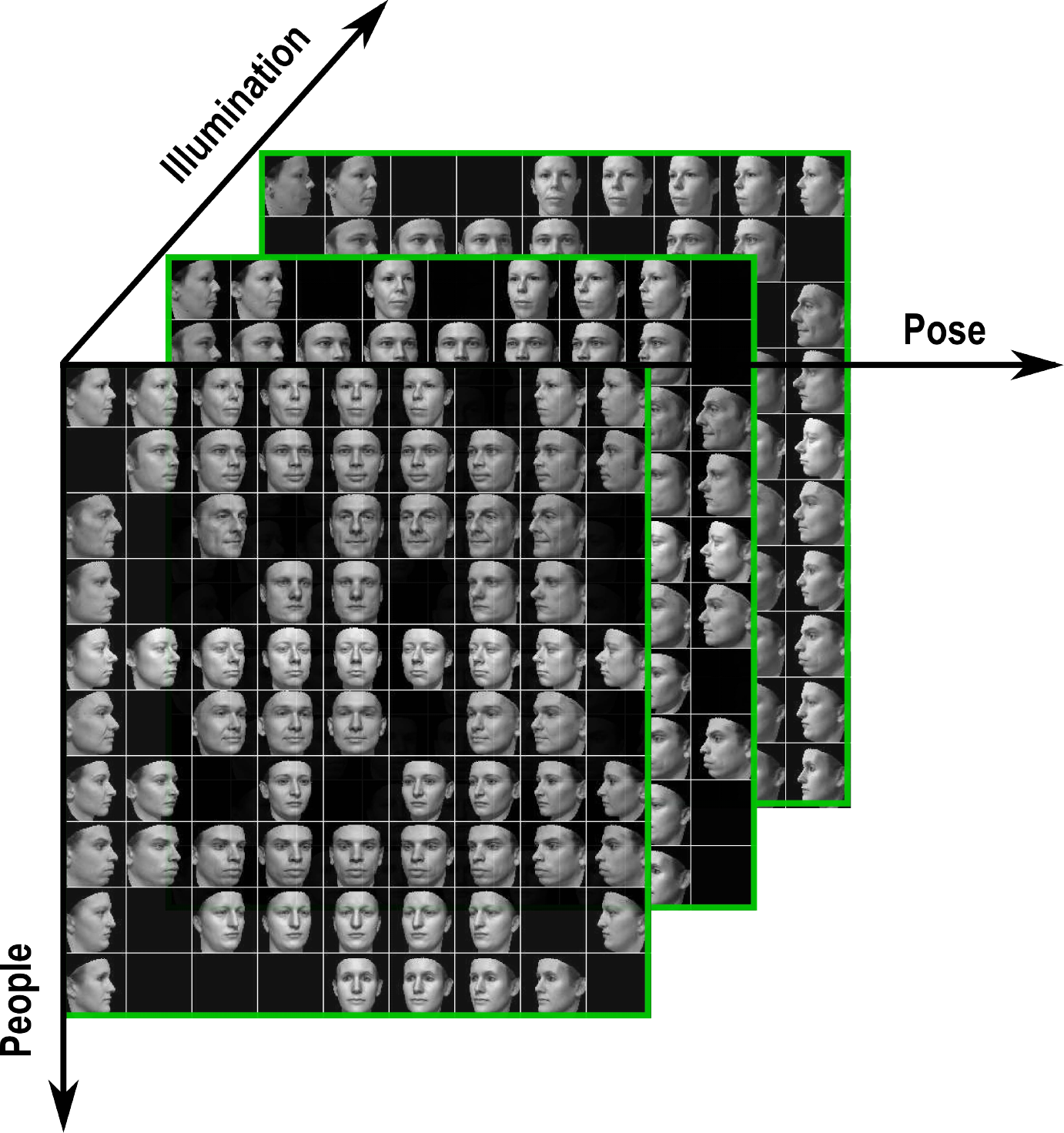}
  \caption{Facial images under multiple conditions where some images are fully missing.  }
  \label{fig:faceimages}
  \vspace{-0.1in}
\end{figure}

\begin{figure*}[ht]
\centering
\subfigure[Missing faces]{
   \includegraphics[width=0.23\textwidth] {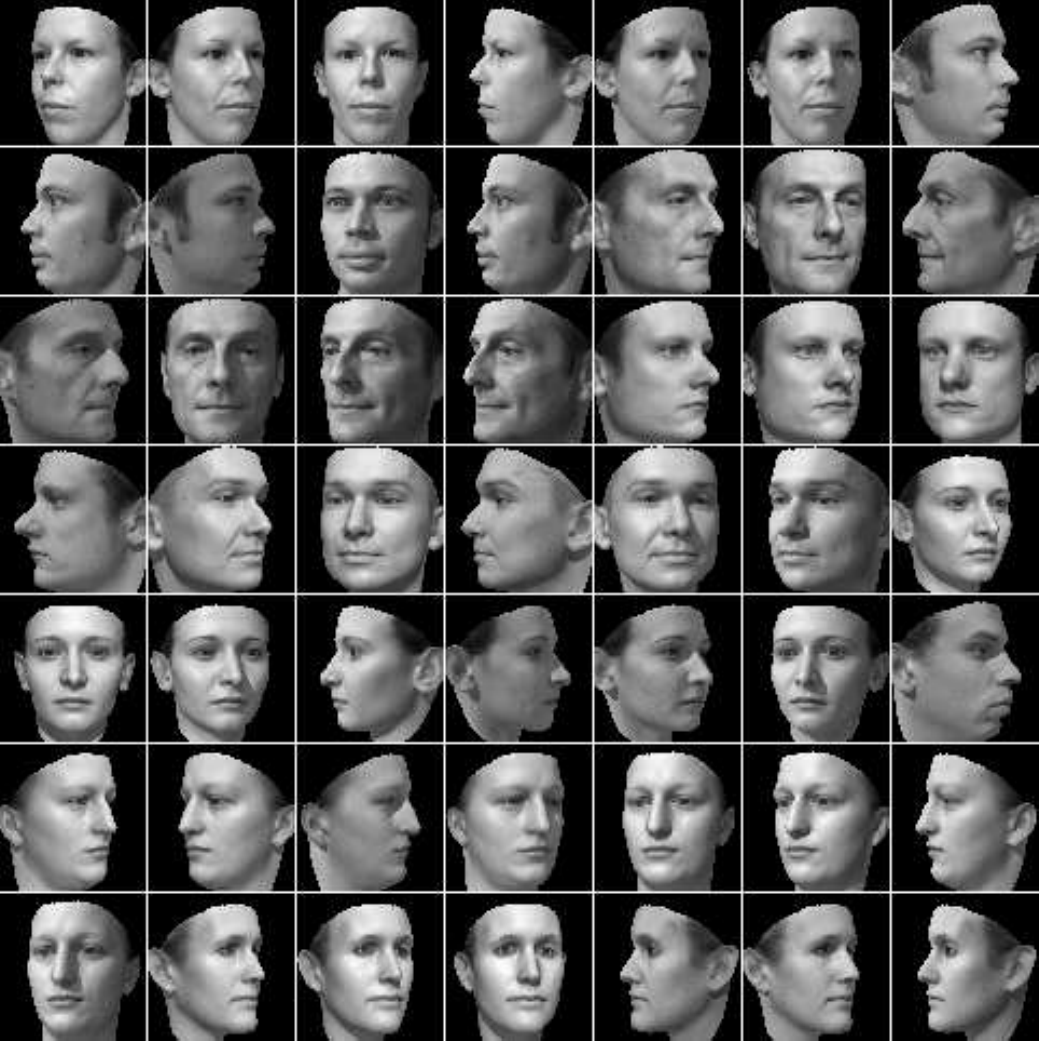}
   \label{fig:missingface-1}
 }
 \subfigure[FBCP]{
   \includegraphics[width=0.23\textwidth] {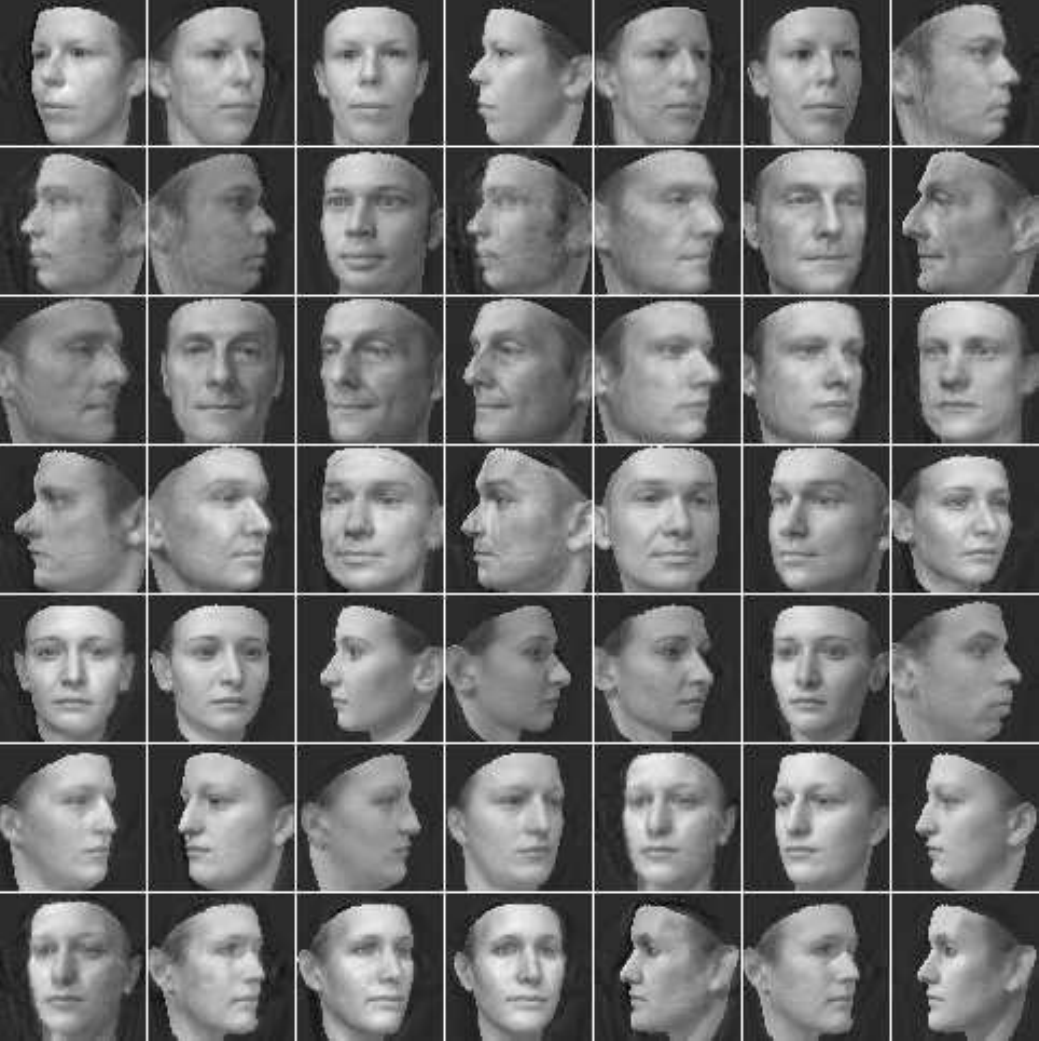}
   \label{fig:missingface-2}
   }
 \subfigure[CPWOPT]{
   \includegraphics[width=0.23\textwidth] {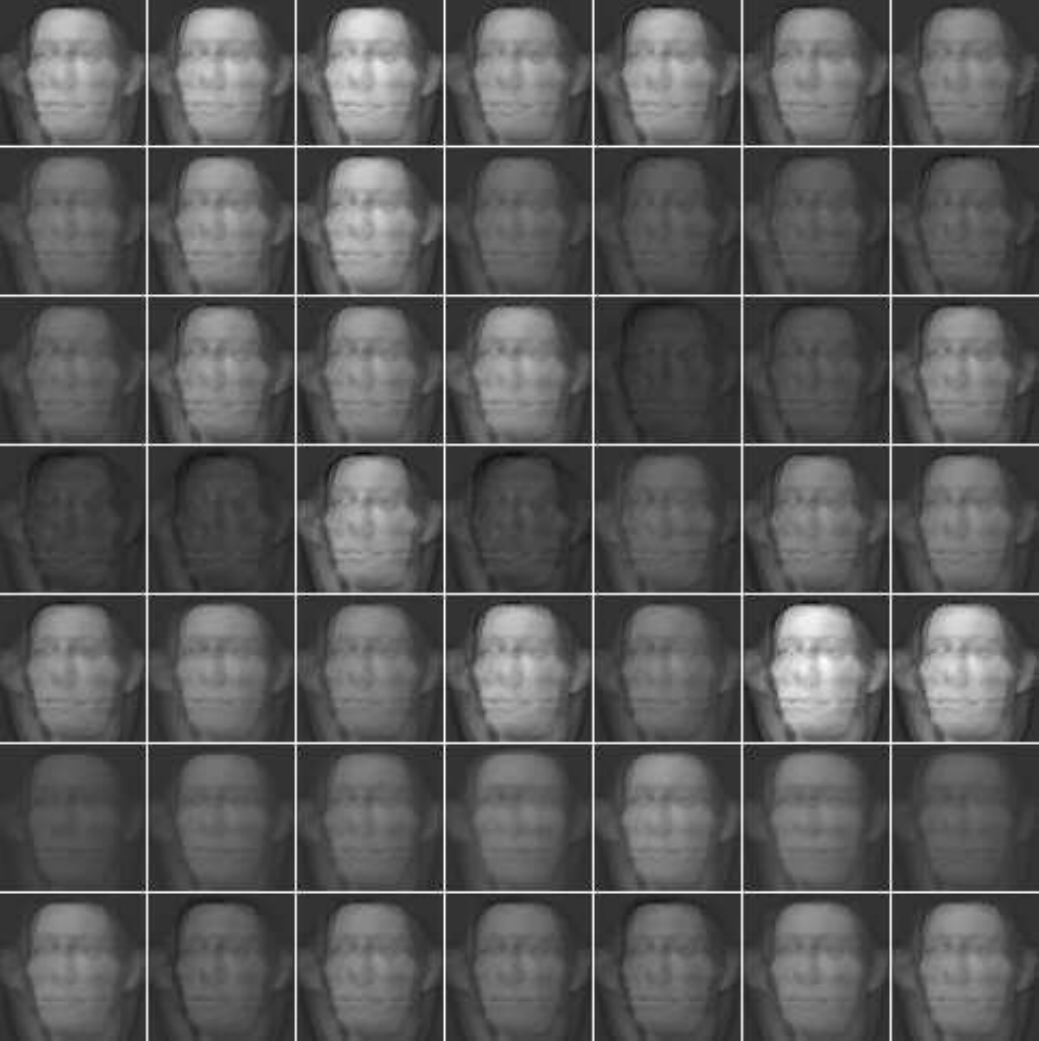}
   \label{fig:missingface-3}
   }
   \subfigure[HaLRTC]{
   \includegraphics[width=0.23\textwidth] {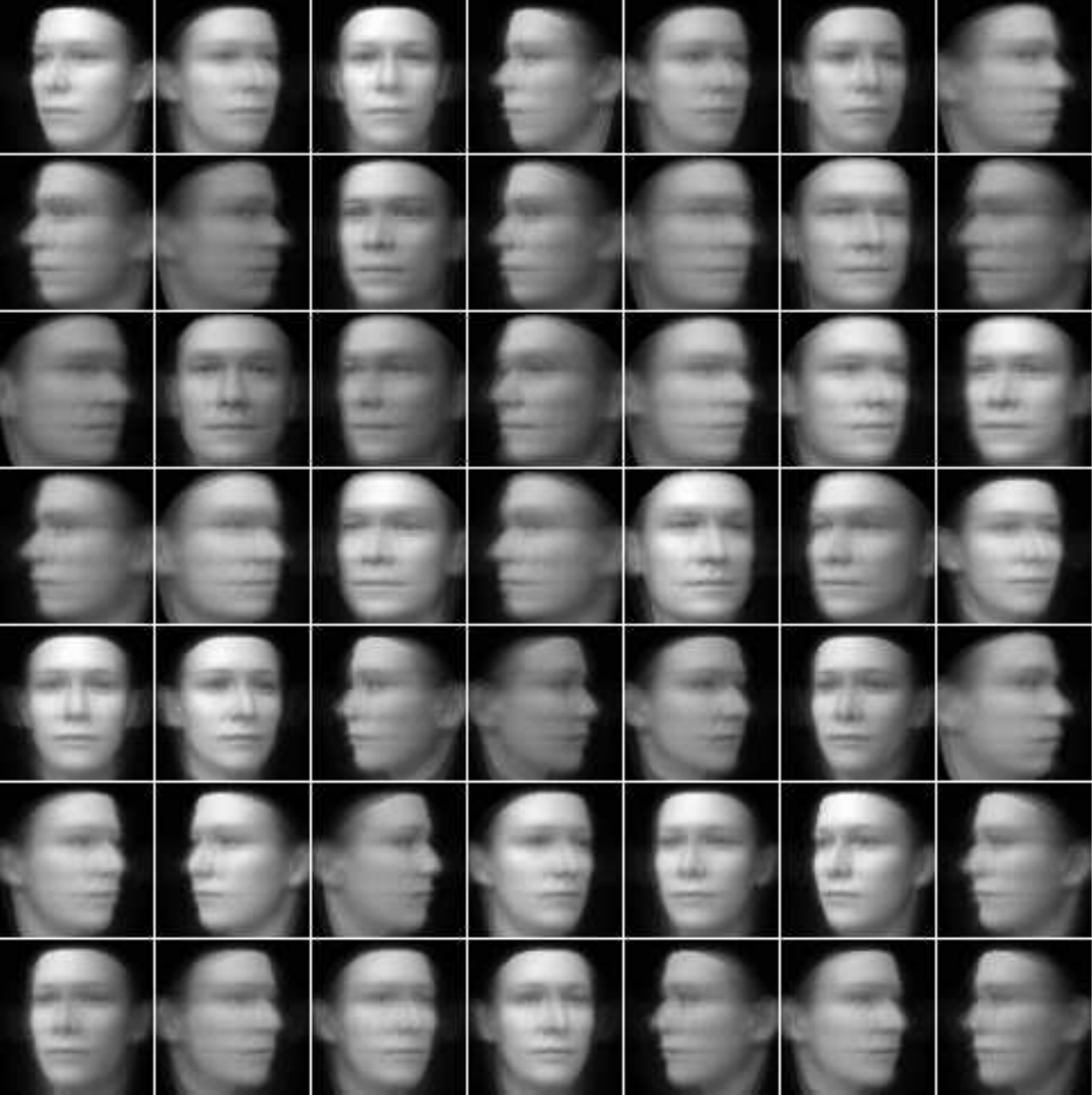}
   \label{fig:missingface-4}
 }
\caption{The ground-truth of 49 missing facial images and the synthetic images by different methods. }
\label{fig:3dfaceMissing}
\end{figure*}

We used the dataset of 3D Basel Face Model~\cite{paysan20093d}, which contains an ensemble of facial images of 10 people, each rendered in 9 different poses under 3 different illuminations.  All 270 facial images were decimated and cropped to $68\times 68$ pixels, and were then represented by a fourth-order tensor of size $4624 \times 10 \times 9 \times 3$. As shown in Fig.~\ref{fig:faceimages}, some images were fully missing. Since some methods are either computationally intractable or not applicable to \mbox{$N\geq4$} order tensor, five algorithms were finally applied on this dataset under different missing ratios. The initial rank was set to 100 in factorization based methods, while the parameters of completion based methods were well tuned based on the ground-truth of missing images.

As shown in Fig.~\ref{fig:3dfaceMissing}, the visual effects of image synthesis produced by FBCP are significantly superior to those produced by other methods. Although both CPWOPT and HaLRTC produce images that are too smooth and blurred, HaLRTC obtains much better visual quality than CPWOPT. The detailed performances are compared in Table~\ref{tab:FaceCompletion}, where RSE w.r.t. observed entries reflects the performance of model fitting, and RSE w.r.t. missing entries particularly reflects the predictive ability. Note that RSE $=$N/A implies that HaLRTC and FaLRTC donot model the observed entries. The inferred rank by FBCP is within the range of \mbox{[98, 100]} that are close to the initialization. Observe that completion based methods including HaLRTC, FaLRTC and HardC. achieve better performance than CPWOPT. However, FBCP demonstrates the possibility that factorization-based scheme can significantly outperform completion-based methods, especially in terms of performance on missing images.

\begin{table}
\renewcommand{\arraystretch}{1.2}
\caption{\small RSEs on observed images (O) and missing images (M). The cases of 36, 49, 64 and 81 missing images were tested. }
\label{tab:FaceCompletion}
\centering
\resizebox{0.48\textwidth}{!}
{
\begin{tabular}{c c c  c c  c c c c}
\hline\hline
\multirow{2}{*}{Method} &  \multicolumn{2}{c}{36/270} & \multicolumn{2}{c}{49/270} & \multicolumn{2}{c}{64/270}  & \multicolumn{2}{c}{81/270} \\
\cmidrule(lr){2-3} \cmidrule(lr){4-5} \cmidrule(lr){6-7} \cmidrule(lr){8-9}
        & O      & M      & O     & M      & O      & M     & O      & M \\
\hline
FBCP    & 0.05    &\bf 0.10    & 0.05    &\bf 0.10    & 0.05    &\bf 0.15   & 0.05    &\bf 0.20\\
CPWOPT  &0.50    &0.65    &0.55    &0.61    &0.57    &0.59   &0.62    &0.73\\
HaLRTC  & N/A       &0.28    &N/A       &0.30    &N/A    &0.31   &N/A    &0.34\\
FaLRTC  & N/A   &0.28    &N/A   &0.30    &N/A    &0.31   &N/A    &0.34\\
HardC.  &0.37    &0.37    &0.36    &0.40    &0.36    &0.40   &0.36    &0.40\\
\hline\hline
\end{tabular}
}
\vspace{-0.1in}
\end{table}

\section{Conclusion}
\label{sec:conclusions}
In this paper, we proposed a fully Bayesian CP factorization which can naturally handle incomplete and noisy tensor data. By employing hierarchical priors over all unknown parameters, we derived a deterministic model inference under a fully Bayesian treatment. The most significant advantage is automatic determination of \emph{CP} rank. Moreover, as a tuning parameter-free approach, our method avoids the parameter selection problem and can also effectively prevent overfitting. In addition, we proposed a variant of our method by using mixture priors, which shows advantages on natural images with a large amount of missing pixels. Empirical results validate the effectiveness in terms of discovering the ground-truth of tensor rank and imputing missing values for an extremely sparse tensor. Several real-world applications, such as image completion and image synthesis, demonstrated the superiority of our method over state-of-the-art techniques. Due to several interesting properties, our method would be attractive for many potential applications.

\bibliographystyle{IEEEtran}
\bibliography{IEEEabrv,BayesianTensor}

\begin{IEEEbiography}[{\includegraphics[width=1in,height=1.25in,clip,keepaspectratio]{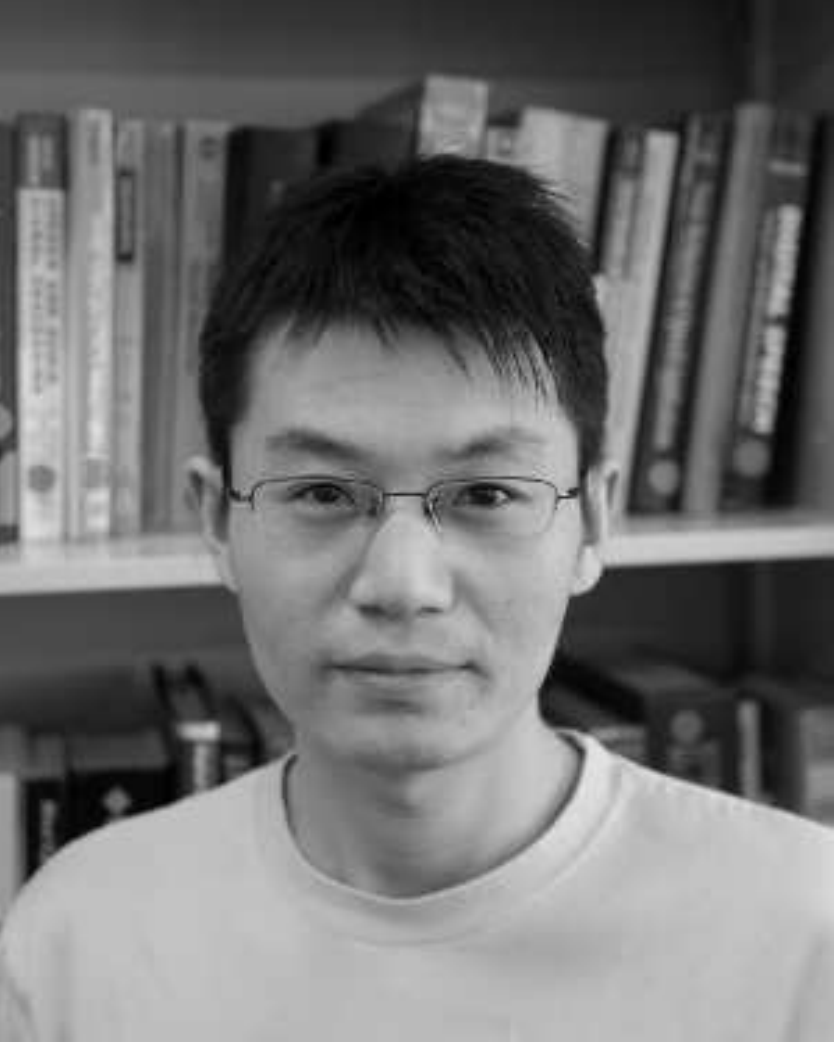}}]{Qibin Zhao} received the Ph.D. degree in engineering from Department of Computer Science and Engineering, Shanghai Jiao Tong University, Shanghai, China, in 2009. He is currently a research scientist at Laboratory for Advanced Brain Signal Processing in RIKEN Brain Science Institute, Japan. His research interests include machine learning, tensor factorization, computer vision and brain computer interface.
\end{IEEEbiography}
\vspace{-0.1in}
\begin{IEEEbiography}[{\includegraphics[width=1in,height=1.25in,clip,keepaspectratio]{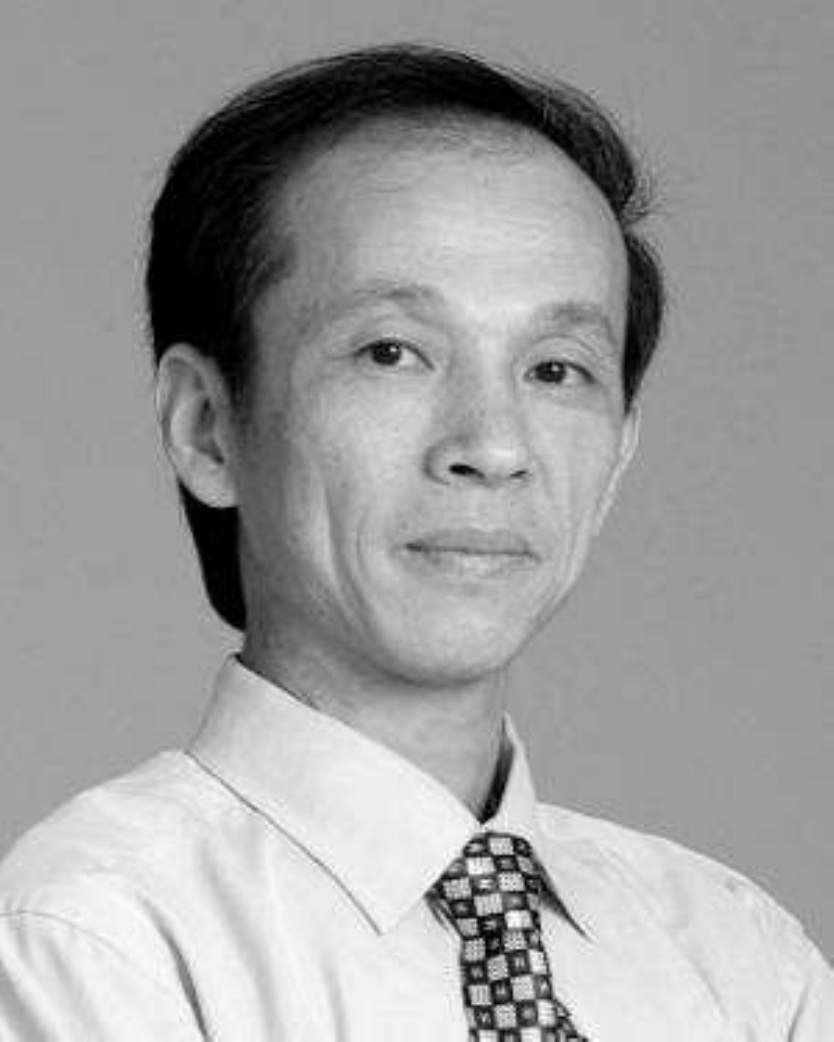}}]{Liqing Zhang} received the Ph.D. degree from Zhongshan University, Guangzhou, China, in 1988.  He is now a Professor with Department of Computer Science and Engineering, Shanghai Jiao Tong University, Shanghai, China. His current research interests cover computational theory for cortical networks, brain-computer interface, perception and cognition computing model, statistical learning and inference. He has published more than 170 papers in international journals and conferences.
\end{IEEEbiography}
\vspace{-0.1in}
\begin{IEEEbiography}[{\includegraphics[width=1in,height=1.25in,clip,keepaspectratio]{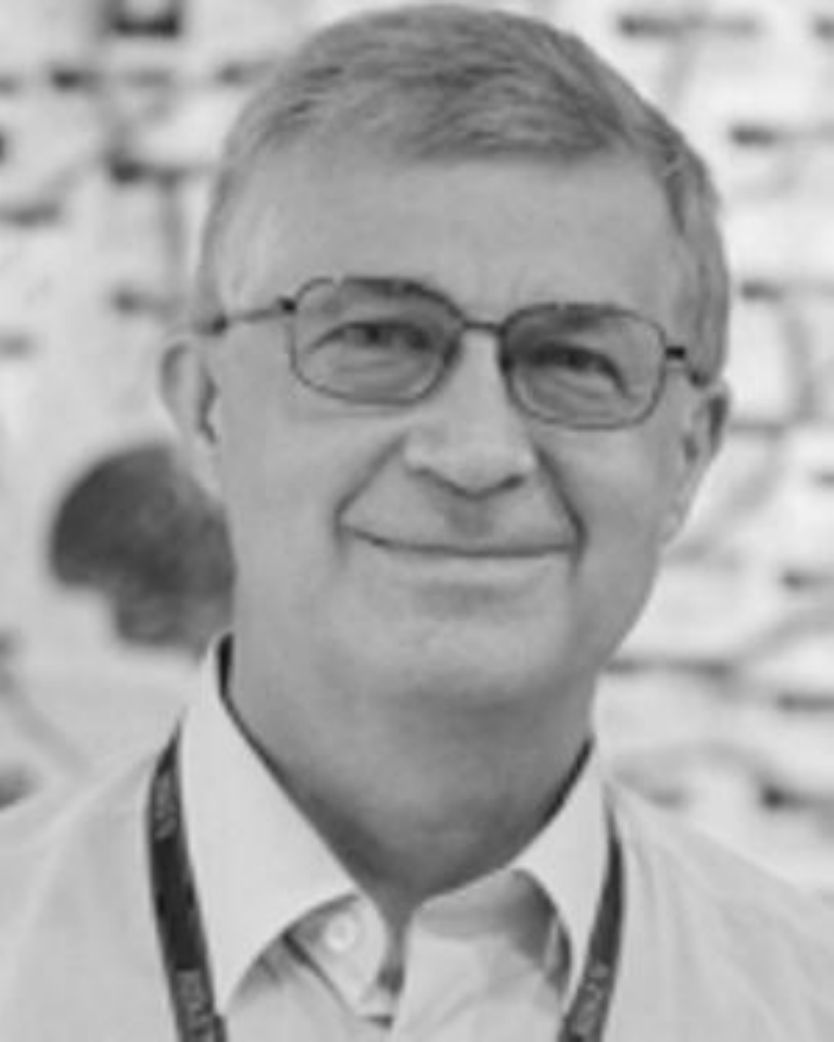}}]{Andrzej Cichocki} received the Ph.D. and Dr.Sc. (Habilitation) degrees, all in electrical engineering, from Warsaw University of Technology (Poland). He is currently the senior team leader of the laboratory for Advanced Brain Signal Processing, at RIKEN Brain Science Institute (JAPAN). He is co-author of more than 250 technical papers and 4 monographs (two of them translated to Chinese). He is  Associate Editor of Journal  of  Neuroscience Methods and  IEEE Transaction on Signal Processing.
\end{IEEEbiography}
\vfill\vfill\vfill\vfill\vfill\vfill\vfill
\end{document}